\documentclass{ieeeaccess}
\usepackage{cite}
\usepackage{amsmath,amssymb,amsfonts}
\usepackage{algorithmic}
\usepackage{graphicx}
\usepackage{textcomp}
\usepackage{multirow}
\usepackage{makecell}
\def\BibTeX{{\rm B\kern-.05em{\sc i\kern-.025em b}\kern-.08em
    T\kern-.1667em\lower.7ex\hbox{E}\kern-.125emX}}
\begin{document}
\history{Date of publication xxxx 00, 0000, date of current version xxxx 00, 0000.}
\doi{10.1109/ACCESS.2023.0322000}

\title{Robust Representation of Oil Wells' Intervals via Sparse Attention Mechanism}
\author{\uppercase{Alina Ermilova}\authorrefmark{1},
Nikita Baramiia\authorrefmark{1}, 
Valerii Kornilov\authorrefmark{1}, 
Sergey Petrakov\authorrefmark{1}, \\
and Alexey Zaytsev\authorrefmark{1, 2}}

\address[1]{Skolkovo Institute of Science and Technology, Bolshoy Boulevard 30, bld. 1, 121205 Moscow, Russia}
\address[2]{BIMSA, 11th Building, Yanqi island, Huairou district, Beijing, China (e-mail: A.Zaytsev@skoltech.ru)}
\tfootnote{This work was supported in part by Grant No. 70-2021-00145 02.11.2021.}

\markboth
{Ermilova \headeretal: Robust Representations of Oil Wells' Intervals via Sparse Attention Mechanism}
{Ermilova \headeretal: Robust Representations of Oil Wells' Intervals via Sparse Attention Mechanism}

\corresp{Corresponding author: Alina Ermilova (e-mail: A.Rogulina@skoltech.ru).}

\begin{abstract}
Transformer-based neural network architectures achieve state-of-the-art results in different domains, from natural language processing (NLP) to computer vision (CV). 
The key idea of Transformers, the attention mechanism, has already led to significant breakthroughs in many areas. 
The attention has found their implementation for time series data as well. However, due to the quadratic complexity of the attention calculation regarding input sequence length, the application of Transformers is limited by high resource demands. Moreover, their modifications for industrial time series need to be robust to missing or noised values, which complicates the expansion of the horizon of their application. 

To cope with these issues, we introduce the class of efficient Transformers named \textbf{Regu}larized Trans\textbf{former}s (Reguformers). 
We implement the regularization technique inspired by the dropout ideas to improve robustness and reduce computational expenses. 
The focus in our experiments is on oil\&gas data, namely, well logs, a prominent example of multivariate time series. The goal is to solve the problems of similarity and representation learning for them.  
To evaluate our models for such problems, we work with an industry-scale open dataset consisting of well logs of more than $20$ wells.    
The experiments show that all variations of Reguformers outperform the previously developed recurrent neural networks (RNNs), classical Transformer model, and robust modifications of it like Informer and Performer in terms of well-intervals' classification and the quality of the obtained well-intervals' representations. Moreover, the sustainability to missing and incorrect data in our models exceeds that of others by a significant margin.
The best result that the Reguformer achieves on well-interval similarity task is the mean PR~AUC score equal to $0.983$, which is comparable to the classical Transformer with PR~AUC equal to $0.984$ and outperforms the previous LSTM model ($0.951$) and two recent the most promising modifications Informer ($0.974$) and Performer ($0.97$).

\end{abstract}


\titlepgskip=-21pt

\maketitle

\section{Introduction}
\label{sec:introduction}
\PARstart{D}{rilling} wells is a time- and money-consuming process that plays an essential role in the exploration of a basin and its characteristic and prevention of drilling accidents \cite{9438672}.
Well and well-interval similarity can help to make decisions about drilling reasonableness: it helps to reconstruct the properties of an oil well by comparing it to other wells with known properties. 
Moreover, with a proper similarity learning approach, we can obtain low-dimensional representations of well-intervals that can help to estimate their lithological and physical properties by having an appropriate well-interval representation of known wells and a new one. 

Many methods to estimate well and well-interval similarity exist as described in Section \ref{sec:related_work}, among which approaches dealing with sequential structures of data seem the most promising. During drilling, engineers collect data along the well in a sequential manner, producing a multivariate sequence of observations along the well depth. 
The sequential nature of data tempts us to use an architecture that exploits its structure for maximum benefit. 
Recurrent Neural Networks (RNNs) have previously been used to obtain representations of well intervals \cite{romanenkova2022similarity}. Despite the widespread use of RNNs, the attention mechanism has made a significant breakthrough in the field of natural language processing (NLP), leading to the creation of the Transformer architecture, which avoids the common problems of RNNs  \cite{vaswani2017attention}. The most evident benefit is that it allows a long-term memory for our model and the equal treatment of whole sequences without focusing on the end of a sequence inherent to RNNs  \cite{ismail2019input}.

However, the processing of long sequences by Transformers requires significant computational time and memory resources and is prone to errors if the quality of input data is low \cite{morvan2022dont, bagla2021noisy}.
Recent approaches  \cite{tay2022efficient}, such as Informer  \cite{zhou2021informer} and Performer  \cite{choromanski2020rethinking}, address this problem by proposing several improvements to the basic Transformer architecture, allowing to focus attention only on the relevant part of a sequence, the more details are provided in Section \ref{sec:methods}.
Nevertheless, the question about the robustness of all the existing models during time series processing remains debatable. There are some papers claiming the ineffectiveness and the low quality of Transformers, e.g., the authors of \cite{zeng2023transformers} show that simple linear-based models outperform Transformers in time series forecasting. 
Nonetheless, the authors of \cite{nie2022time} propose Transformers' adoption for time series data, which includes patch strategy and the usage of the Transformer's backbone. This allows to outperform simple linear models proposed in \cite{zeng2023transformers}. 
So, these articles confirm that Transformers may perform poorly in some tasks, which also highlights the need for robust models.
In addition, in our problem statement, the expert's labeling is not required as we can formulate the task in the way of well-interval similarity instead of well similarity, which leads to the self-supervised nature of our method.  

Given all this, we present our approach to similarity learning for well-logging data based on the Transformer architecture. The main contributions of this work are as follows:
\begin{enumerate}
    \item We propose a new variant of the Transformer architecture based on the regularization, \textit{Reguformer}. This model adopts the dropout ideas to create more accurate, efficient and robust models for processing of time series data.
    \item The basic idea of these models, dropout, makes the models both more efficient and more robust, which leads to the increase of the Reguformer's performance against missing or noised data.
    \item As a self-supervised approach is used, expert labels are not needed during our model training and can be applied only for fine-tuning if required. This architecture improves existing results for time series similarity estimation and representation learning tasks.
    \item We implement all modifications of Reguformer to the oil\&gas data and solve the problem of similarity estimation between well-intervals. Our models significantly improve existing results, proving the usefulness of Reguformers for processing logging data for oil\&gas wells. The evidence comes from a better similarity estimation and overall better representations. 
\end{enumerate}

\section{Related work}
\label{sec:related_work}
To compare wells, we can consider well logging information. It is represented as sequential data of petrophysical numerical properties recorded by lowering a variety of sensors into an oil well.

The simplest way to estimate the correlation between wells is a rule-based approach. The authors of  \cite{startzman1987rule} rely on prior expert knowledge to identify logical rules for the well-to-well correlation. In  \cite{zoraster2004curve} and  \cite{ali2021machine}, the use of geometric distances and Jaccard and Overlap similarities are proposed for wells' similarity calculation. The authors of \cite{6809843} adopt synchronization likelihood and visibility graph similarity for well similarity assessment. However, most of these methods consider oversimplified physical well properties and are unproductive in terms of obtaining well or well-interval representations. Moreover, they work with only two features. It is reasonable to assume that using a larger number of features will reveal more complex data dependencies, as well logs consist of many characteristics.

The experiments presented in  \cite{rogulina2022similarity} show better performance of classical machine learning models than rule-based approaches. The model based on gradient boosting shows a sufficiently high quality with PR~AUC~$0.943$. Although the predictions can be used for distance calculation and clustering, representations reflecting the geographical and physical characteristics of the wells are not possible in this case due to the model's architecture.

Deep neural networks are useful for generating well-interval representations (embeddings). The authors of  \cite{romanenkova2022similarity} obtain well-interval representations by LSTM. In  \cite{kumar2021transformer}, the authors moved forward and implemented a transformer-based approach to well logs processing. This approach is implemented for well diagnosis. In  \cite{egorov2022self}, autoencoder-based approaches are considered to provide data representations applicable to different oil\&gas field problems. The authors of  \cite{marusov2022non} adapt non-contrastive approaches like BYOL and Barlow Twins for logging data representations. In  \cite{abdrakhmanov2021development}, the authors train transformers to predict wells' productivity. It is claimed that this approach allows the model to be transferred to another well to get an accuracy increase in the bottomhole pressure or flow rate evolution for each well prediction. 
There also studies devoted to the adoption of CNN \cite{abad2021predicting} and autoencoders \cite{alakeely2021application} for oil flow rate prediction.

As mentioned above, well logs can be considered sequential data, specifically, time series. The classical approaches to estimating the similarity without additional information about separate time series come from unsupervised learning. 
Among all the models in this area, autoencoders  \cite{rumelhart1986learning} seem the most promising in terms of learning data representations. Another popular method, as stated before, is RNNs such as GRU  \cite{cho2014learning}, and LSTM  \cite{hochreiter1997long}. However, all unsupervised learning methods suffer from their results' inaccuracy because of the absence of labels. 
Due to the drawbacks, self-supervised models show better performance, producing an outcome that is more aligned with the current task. The two main ideas of self-supervised learning are the following: 
\begin{enumerate}
    \item \underline{Generative}. The most prominent representative is autoregression  \cite{klein1997statistical}, which predicts the future using representations. Autoencoders considered above can also be used in the context of self-supervised learning. 
    \item \underline{Contrastive}. The key idea of these approaches is that similar objects should have similar representations, and dissimilar objects should have distant representations. The most common loss functions in this class are Siamese and Triplet.
\end{enumerate}

In this paper, we focus on contrastive learning approaches. 

Within Transformer-based models, the attention mechanism is associated with the highest costs of memory as well as computational quadratic complexity. This is due to the need to store and make operations with all the elements of the attention matrix, which is quite large. It is noticeable in our case as we deal with long well-log sequences. Much research is devoted to reducing these difficulties, mainly two approaches. First, attention matrix sparsification, for example,  Informer  \cite{zhou2021informer}, Performer  \cite{choromanski2020rethinking},
Linformer  \cite{wang2020linformer}, Longformer  \cite{beltagy2020longformer},  BigBird  \cite{zaheer2020big}. Second, bucket processing application, for instance, Reformer  \cite{kitaev2020reformer}. 
These simplifications allow for obtaining linear computational complexity. In this paper, we also propose the Regularized Transformer (Reguformer) class, adopt them for solving well-interval similarity task, and compare different regularization techniques among which the top queries of Informer is present with the classical Transformer and another efficient modification Performer.  
More details are available in the recent review  \cite{tay2022efficient}, and some performance comparisons can be found in~\cite{tay2020long}.

The interpretability of the Transformer-based model is closely tied to the basis of this model, which is the mechanism of attention  \cite{chefer2021transformer}, as the attention matrix provides deep insight into the object and relations in it  \cite{cherniavskii2022acceptability}. Moreover, the attention exposes the main contribution behind each specific output generated by the model. In particular, scores of attention to input areas or intermediate objects are interpreted as a measure of the contribution of the visited element to the output of the model  \cite{hao2021self}. 
This idea allows us to overcome the difficulties associated with the fact that RNNs such as LSTM have a bottleneck related to the concentration of all information in a single vector that follows the encoder in sequence to sequence models  \cite{ismail2019input}.

Uniting everything together, we see that these methods have not been developed and applied to well-interval similarity problems due to challenges related to long time intervals and the low amount of data used. 
We expect to fill this gap by proposing an efficient and interpretable attention-based model trained in a self-supervised manner.

\section{Methods}
\label{sec:methods}

\subsection{Data Overview}
\label{subsec:data_overview}

We consider an open access dataset provided by the New Zealand Petroleum \& Minerals Online Exploration Database  \cite{web:newzeland1}, and the Petlab  \cite{web:newzeland2}. It consists of wells from the Taranaki basin, New Zealand. We use the data from the largest formation Urenui. 

In our work, we consider similarity not between whole wells but between intervals from wells of length $100$ measurement or $33$ ft. Such selection allows appropriate aggregation of the local properties of a rock. 

In all our experiments, we use the following four features: porosity inferred from density
log (DRHO), density log (DENS), gamma-ray (GR), and  sonic log (DTC).

\subsection{Methodology}
\label{subsec:methodology}

\subsubsection{Similarity problem statement}
\label{subsubsec:problem_statement}

We look at intervals of wells. 
Each interval consists of $4$ features measured during $100$ consecutive points in depth during drilling.
Our goal is to provide a model that can report a similarity between two intervals in terms of their physical properties. 
As exact labeling on similarities is rarely available, we use both direct and indirect approaches to estimate the quality of obtained models.

To train the model, we focus on the well-linking problem  \cite{romanenkova2022similarity}: we consider a pair of intervals as positive if they belong to the same well (target value for this pair is equal to 1) while a pair of intervals is negative if they are from two different wells -- target value, in this case, is equal to 0.
So, we do not require any labels for training for such labeling.
See the problem statement for model training in Figure~\ref{fig:problem_statement}. 
The scope of application of models trained via this approach is wider than distinguishing intervals from different wells, as we hope to obtain representations useful for other problems. 

\Figure[t!](topskip=0pt, botskip=0pt, midskip=0pt)[width=0.99\columnwidth]{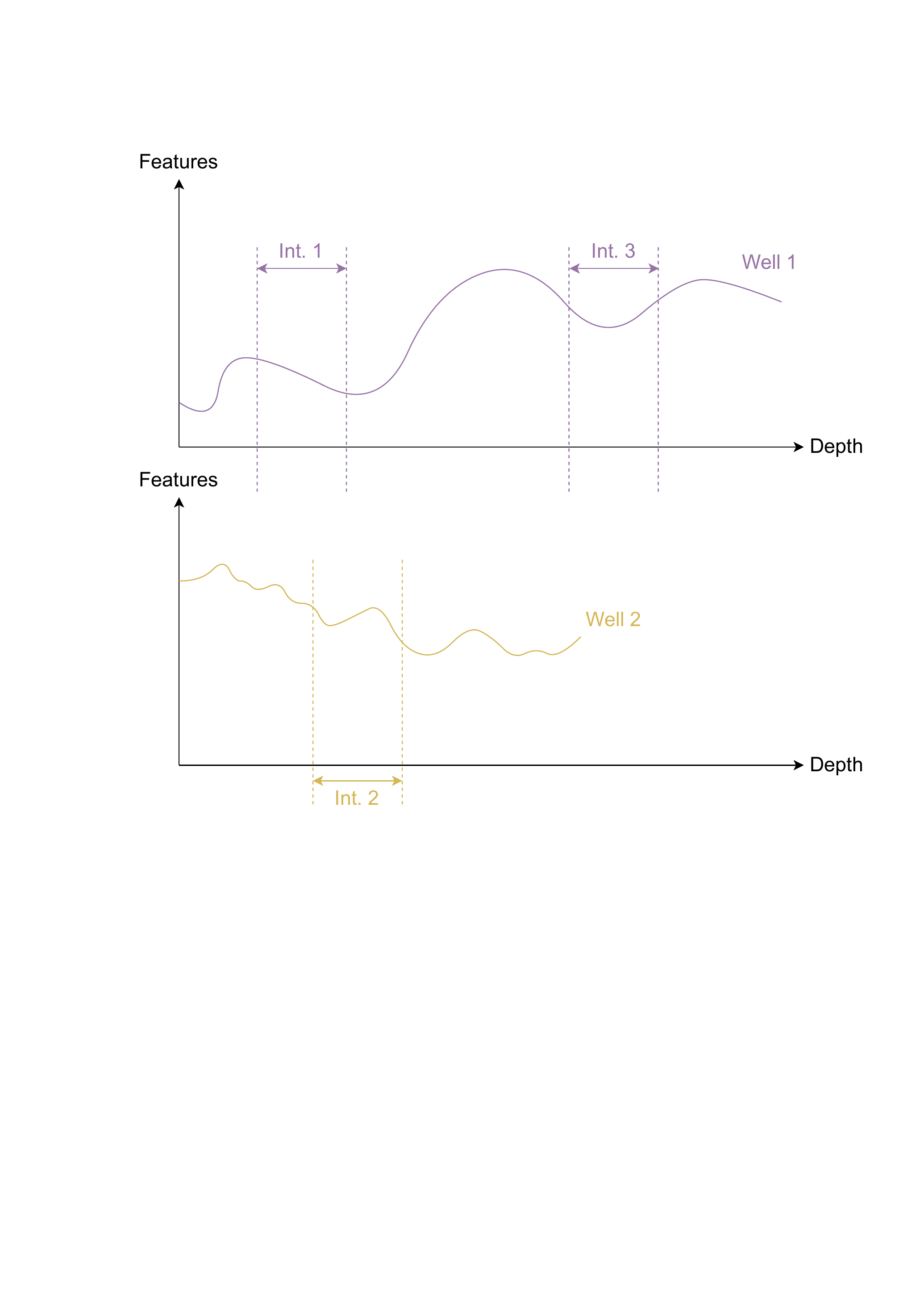}
{ \textbf{Well-linking problem statement. Interval (Int.) 1 and Int. 3 are considered similar, so the prediction for this pair should be as close to one as possible. Intervals 1 and 2 are from different wells, and their similarity prediction should equal zero.}\label{fig:problem_statement}}

We use two loss functions in our experiments: Siamese and Triplet depicted in Figures~\ref{fig:siamese_architecture} and~\ref{fig:triplet_architecture} correspondingly. The Siamese loss requires a pair of intervals with a label indicating whether they belong to the same well or not, while the Triplet loss takes as an input a triplet of intervals: an anchor interval, a positive (an interval from the same well as the anchor), and a negative (an interval belong to the different well as the anchor). 

In our case, the Siamese loss is represented by the binary cross entropy (BCE) between the target and the predictions:

\begin{equation}
BCE = -\frac{1}{n} \left( \sum_{i = 1}^{n}{y_i \log{p_i} + \left( 1 - y_i \right) \log{\left( 1 - p_i \right)}} \right),
\label{siamese_loss}
\end{equation}
where $y_i$ -- target values, $p_i$ -- predicted probabilities. 

For the Triplet loss, we use the standard formula with the Euclidean distance:
\begin{equation}
L = \max \left( ||\mathbf{a}_i - \mathbf{p}_i||_2 - || \mathbf{a}_i - \mathbf{n}_i ||_2 + \text{margin}, 0 \right),
\label{triplet_loss}
\end{equation}
where $\mathbf{a}_i$, $\mathbf{p}_i$, and $\mathbf{n}_i$ are embeddings of anchor, positive,  and negative well-intervals, correspondingly. We consider $\text{margin} = 1.75$. 

\subsubsection{Neural network architectures}
\label{subsubsec:nn_architectures}
As encoders for these architectures, we propose to use different variants of Transformers: basic Transformer, Performer, and Reguformer with different regularization strategies.
Details on the architectures are provided below.

\Figure[t!](topskip=0pt, botskip=0pt, midskip=0pt)[width=0.99\columnwidth]{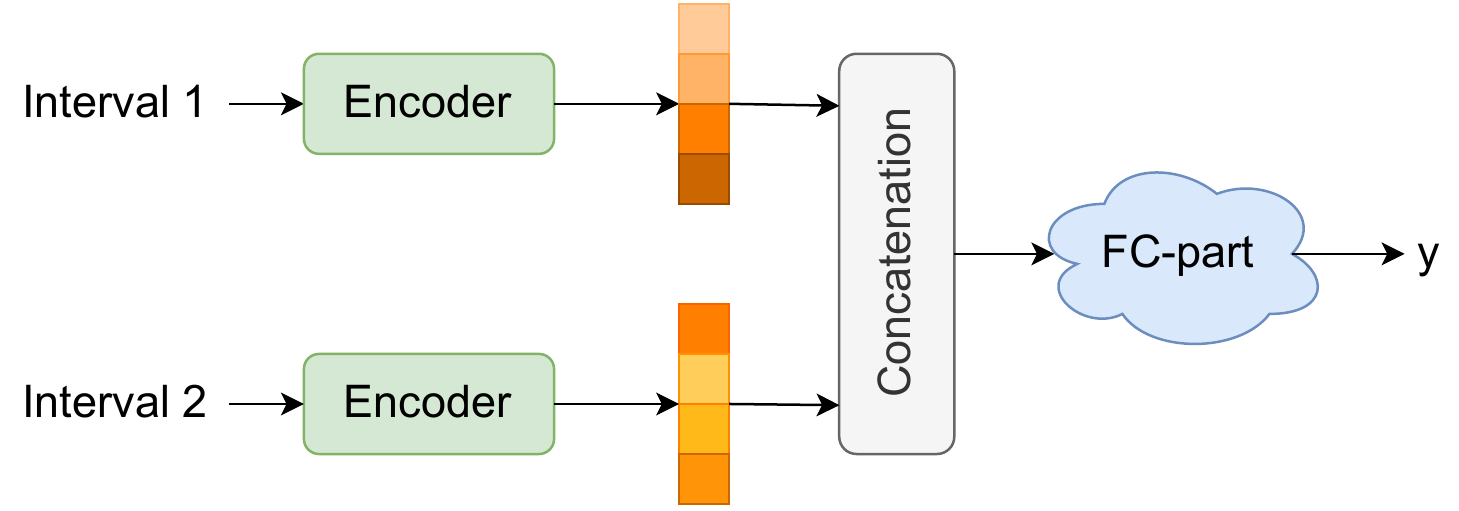}
{ \textbf{Siamese loss function. It can use an additional fully-connected part (FC-part) to predict the target similarity for a pair of intervals.}\label{fig:siamese_architecture}}

\Figure[t!](topskip=0pt, botskip=0pt, midskip=0pt)[width=0.99\columnwidth]{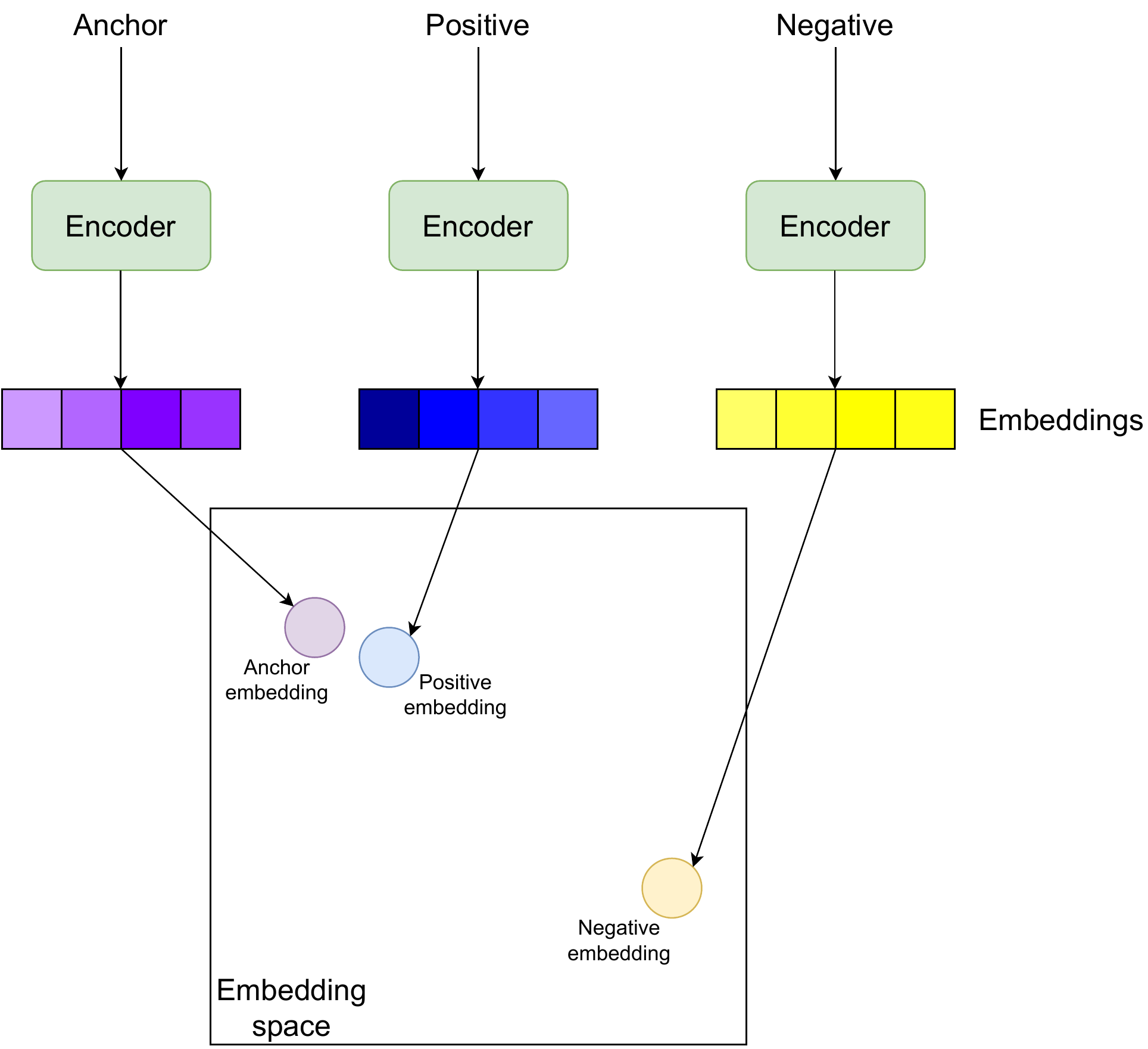}
{ \textbf{Triplet architecture with the desired distribution of embeddings in the embedding space.}\label{fig:triplet_architecture}}

\paragraph{Transformer}
\label{transformer}
It is a neural network architecture introduced for NLP tasks  \cite{vaswani2017attention}.
It outperforms alternative recurrent and convolution architectures in many problems. Besides recursion avoidance that speeds up calculations via parallel computation during both training and inference steps, the authors introduced a self-attention mechanism in the encoder part, which we utilize as a backbone for a classification task. 
 
The model input $X$ is a sequence of length $L$ described by the $f$ features ($Q \in {\mathbb{R}^{L \times f}}$). Transformer processes it via $N$ layers: input to the first layer is $X$. Inputs to the following layers are outputs of corresponding previous layers. The input sequence length is $L$ for all layers. Each layer consists of an attention block and position-wise feed-forward block (or feed-forward network, FFN) and is followed by layer normalization with skip connection.

The attention block is a Query-Key-Value (QKV) model $\mathcal{A} (Q, K, V)$, which helps to use connections between points in a sequence:
\begin{equation}
\mathcal{A} \left(Q, K, V\right) = \mathrm{softmax} \left(Q K^T / \sqrt{d} \right) V, 
\label{transformer_attention}
\end{equation}
where $Q \in {\mathbb{R}^{L \times d}}$ -- query matrix, $K \in {\mathbb{R}^{L \times d}}$ -- key matrix, $V \in {\mathbb{R}^{L \times d}}$ -- value matrix. 
In encoder $Q = X W^Q$, $K = X W^K$, $V = X W^V$ where $X$ is the output from the previous layer.
Matrix multiplication of $Q$ and $K$ with softmax provides the attention matrix $A$. It consists of weights for $V$ matrix elements that share information between different words (or log data from different well levels).
Thus, the elements of the attention matrix provide information about correspondence between different parts of a sequence.
Higher attention results from closer correspondence between a key and a query, and, thus, reflect closer correspondence between different elements of a sequence.
The division by $d$ is used to mitigate the vanishing gradient problem.
We can reformulate the formula for the attention of the $i^{\text{th}}$ query calculation. According to \cite{tsai2019transformer}, a
kernel smoother in a probability form define the $i^{\text{th}}$ query attention \cite{zhou2021informer}: 
\begin{equation}
\mathcal{A} \left(\mathbf{q}_i, K, V\right) = \sum_j{\frac{k\left(\mathbf{q}_i, \mathbf{k}_j\right)}{\sum_l{k\left(\mathbf{q}_i, \mathbf{k}_l\right)}} \mathbf{v}_j} = \sum_j{p\left(\mathbf{k}_j | \mathbf{q}_i \right)} , 
\label{transformer_attention_i_query_format}
\end{equation}
where $k(\mathbf{q}_i, \mathbf{k}_j) = \exp{\left( \frac{\mathbf{q}_i \mathbf{k}_j^T}{\sqrt{d}} \right)}$ is the asymmetric exponential kernel and $p\left(\mathbf{k}_j | \mathbf{q}_i \right) = \frac{k(\mathbf{q}_i, \mathbf{k}_j)}{\sum_l{k\left(\mathbf{q}_i, \mathbf{k}_l\right)}}$. 

However, it is beneficial to utilize not one attention function but several with different projection matrices to retrieve different information from data, so MultiHead attention was introduced:
\begin{equation}
\mathcal{MA} (Q, K, V) = \mathrm{concat} (\mathrm{head_1, \dots, head_H}) W^O,
\label{multihead_attention}
\end{equation}
where $\mathrm{head_i} = \mathcal{A} (X W_i^Q, X W_i^K, X W_i^V)$.

Processing the sequence as a whole avoids problems of forgetting past information, which is typical for alternative architecture RNNs. 

The position-wise FFN is a fully-connected neural network, which is applied to each element in sequence separately and identically:
\begin{equation}
\mathrm{FFN}(x) = \max (0, x W_1 + b_1) W_2 + b_2.
\label{transformer_ffn}
\end{equation}

\paragraph{\underline{Regu}larized Trans\underline{former} (Reguformer)} Inspired by the dropout technique, we propose a generalized class of models: Regularized Transformers, which consists of $8$ models. All these models inherit the attention mechanism from the Transformer but implement it more efficiently in terms of memory consumption and computational cost. 

For all the Reguformers, the attention is calculated in the following way:
\begin{equation}
\mathcal{A} (Q, K, V) = \mathrm{softmax} \left( \bar{Q} \bar{K}^\top / \sqrt{d} \right) V,
\label{reguformer_attention}
\end{equation}

The idea is to use only:
\begin{itemize}
    \item random queries ($\bar{Q}$ -- random queries, $\bar{K} = K$);
    \item random keys ($\bar{K}$ -- random keys, $\bar{Q} = Q$);
    \item random queries and keys ($\bar{Q}$ -- random queries, $\bar{K}$ -- random keys);
    \item top queries ($\bar{Q}$ -- top queries, $\bar{K} = K$);
    \item top keys ($\bar{K}$ -- top keys, $\bar{Q} = Q$);
    \item top queries and keys ($\bar{Q}$ -- top queries, $\bar{K}$ -- top keys);
    \item random queries and top keys ($\bar{Q}$ -- random queries, $\bar{K}$ -- top keys);
    \item top queries and random keys ($\bar{Q}$ -- top queries, $\bar{K}$ -- random keys).
\end{itemize}

In case of top queries or/and keys, we implement the sparsity measurement proposed in \cite{zhou2021informer}:
\begin{equation}
\mathcal{M} \left(\mathbf{q}_i, K\right) = \log{\sum_j{k(\mathbf{q}_i, \mathbf{k}_j)}} - \frac{1}{L}\sum_j{k(\mathbf{q}_i, \mathbf{k}_j)} , 
\label{informer_sparsity_measure}
\end{equation}
where $k(\mathbf{q}_i, \mathbf{k}_j) = \exp{\left( \frac{\mathbf{q}_i \mathbf{k}_j^T}{\sqrt{d}} \right)}$. 
This method is utilized for top queries selection in the Informer architecture \cite{zhou2021informer}. So, Informer is a particular case of Reguformer (when top queries are selected).

\paragraph{Informer} The main novelty of the proposed Informer \cite{zhou2021informer} encoder part is the sparse self-attention mechanism \textit{ProbSparse}. 
The idea is to use only such queries (\ref{transformer_attention_i_query_format}) for which $p\left(\mathbf{k}_j | \mathbf{q}_i\right)$ is far from zero. Consider again $Q$, $K$ and $V$ are query, key and value matrices correspondingly, $\mathbf{q}_i$, $\mathbf{k}_i$, $\mathbf{v}_i$ -- $i^{\text{th}}$ rows of these matrices. Using query sparsity measurement, we can get top queries and define self-attention similar to the original work:
\textit{ProbSparse} self-attention is  
\begin{equation}
\mathcal{A} (Q, K, V) = \mathrm{softmax} \left( \bar{Q} K^\top / \sqrt{d} \right) V,
\label{informer_attention}
\end{equation}
where $\bar{Q}$ is created only from top queries. So, for a single head, we obtain a sparse attention matrix. As we have multiple heads, we avoid severe loss of information via \textit{ProbSparse} self-attention. The original Informer's paper  \cite{zhou2021informer} proposes the sparsity measurement described in \ref{informer_sparsity_measure}. 

Generally, computation complexity for each query-key lookup decreases from quadratic $O(L^2)$ to $O (L \ln L)$, the layer memory usage from $O (L^2)$ to $O (L \ln L)$. 

\paragraph{Performer} The Performer architecture  \cite{choromanski2020rethinking} utilizes \textit{Fast Attention Via positive Orthogonal Random features (FAVOR+)} mechanism that provably achieves linear time and space complexity in $L$ of full-rank attention matrix estimation at any precision. Moreover, it does not rely on prior assumptions about matrix structure.

\textit{FAVOR+} mechanism can be summarised in the following way. Attention matrix $\mathcal{A}$ may be considered a kernel matrix with a kernel $k(x, y) = \exp(x^\top y)$. This allows it to be estimated by unbiased approximation with random feature map $\phi_{trig}$:
\begin{equation}
\mathcal{A} (Q, K, V) = \hat{D}^{-1} (Q^{'}((K^{'})^T V))
\label{performer_attention}
\end{equation}

\begin{equation}
\hat{D} = \mathrm{diag}(Q^{'}((K^{'})^T\textbf{1}_L)
\label{performer_matrix}
\end{equation}
$Q^{'}, K^{'} \in \mathbb{R}^{L \times r}$ have rows $\phi(q_i^T)^T$ and $\phi(k_i^T)^T$ respectively. Random feature maps have to be positive, regularized, and orthogonal to achieve a stable estimation of the attention matrix with low variance. This approach results in exponentially small bounds on large deviation probabilities and gets the algorithm's $O(Ld^2\log d)$ complexity.

There are multiple existing implementations of Performer that can be found online\footnote{\underline{https://github.com/lucidrains/performer-pytorch}}\footnote{\underline{https://github.com/xl402/performer}}\footnote{\underline{https://github.com/nawnoes/pytorch-performer}}. Nevertheless, the theory presented in the paper  \cite{choromanski2020rethinking} may not only be applied to increase the performance of Transformers but also to expand the Transformer architecture class and even go beyond their scope. For example, random feature methods can be used for a broader family of kernels  \cite{peng2021random,munkhoeva2018quadrature}.

\paragraph{Attention analysis.}
To investigate sensitivity to inputs and identify the most important inputs, we can consider the attention matrix~$A$.
We evaluate this hypothesis for our Transformer models in the following ways:
\begin{itemize}
    \item Calculate the correlation between attention weights and models' gradients. According to  \cite{serrano2019attention} and  \cite{jain2019attention}, attention weights seem to be uncorrelated with gradients. Moreover, as stated in  \cite{jain2019attention}, Transformer-based models are resistant to random elimination of parts of an interval, while it is more sensitive to deleting parts with the smallest attention weights. However, in these papers, the authors consider only simple architectures, while the interpretability of the larger model based on attention remains an open question.
    \item Eliminate parts of intervals and examine models' quality. 

    If excluding the most important (according to some criterion) measurements reduces the quality more significantly than a random drop, we can say that the criterion reasonably identifies the most important features. We consider the following criteria for $i$-th element of a sequence: 
    \begin{enumerate}
    \item The smallest attention weights $q_i^\top k_i $. To select top-$k$ queries, we consider the diagonal elements of the attention matrix: we find an index of the $k$ smallest elements and drop the row with this index in the original well-interval. We take the sum of attention matrices over all layers and all heads.    
    \item The highest gradient with respect to the inputs. We select the top-$k$ model's gradients, find their indices, and delete elements in the original well-interval according to these indices.  
    \end{enumerate}
    We use zeros and values from the normal distribution as masks for the masking part when we fill gaps with random values instead of a complete drop of them.
\end{itemize}

\section{Experiments and Results}
\label{sec:experiments_results}

Our experimental evaluation consists of several parts that sequentially answer the following questions: 
\begin{itemize}
    \item Does the usage of models based on transformer architectures improve the similarity models compared to RNNs used previously? 
    \item Does the introduction of the regularization technique increase the models' similarity estimation?
    \item How efficient and robust is the proposed approach? In what way is it better than recurrent architectures?
    \item Can embeddings produced by our Reguformers be reused to solve problems other than similarity estimation?
    \item Does attention add to the interpretability of the model by providing a new way to conduct sensitivity analysis? 
\end{itemize}


In all our experiments, we use the following notation for the regularization techniques of Reguformer:
\begin{equation}
\text{reg\_type}_Q Q\_\text{reg\_type}_K K,
\label{reg_notation}
\end{equation}
where the first part, $\text{reg\_type}_Q Q$, and the second part, $\text{reg\_type}_K K$, stands for the regularization type of query and keys matrices, correspondingly. The two variants for $\text{reg\_type}_Q$ and $\text{reg\_type}_K$ are possible: "top" and "rand", which refers to the top rows selection via sparsity measurement proposed in \cite{zhou2021informer} or random selection, respectively. If the modification includes only query or key matrix regularization, the other part in the regularization notation is skipped. 

The code for all our experiments is available at \texttt{roguLINA/Reguformer} \footnote{\underline{https://github.com/roguLINA/Reguformer}} .

\subsection{Models quality}
\label{subsec:models_quality}
This section compares our models' quality for the well-linking problem. Following the research \cite{romanenkova2022similarity}, for both Siamese and Triplet models, we calculate the Euclidean and cosine distances (Eucl.dist. and Cos.dist.) to obtain the probabilities. In the Siamese case, we also use the $3$ fully-connected layers (FC) with the following exact architecture: \texttt{FC (input\_size, hidden\_size) +
ReLU + Dropout (0.25) + FC (hidden\_size, hidden\_size) + ReLU
+ Dropout (0.25) + FC (hidden\_size, output\_size) + Sigmoid}. More details of the models' training can be found in Section \ref{sec:tech_details}.

We consider the area under the receiver operating characteristic (ROC~AUC) and the precision-recall curve (PR~AUC) to be the most representative metrics. The first one was used during the hyperparameter optimization with Optuna \footnote{\underline{https://github.com/optuna/optuna}}. Figure ~\ref{fig:box_plots_roc_auc} demonstrates the models' quality during the hyperparameter search. Each point represents the model with the exact hyperparameter set. The variants of our Reguformer show comparable scores as the classical Transformer and another efficient modification Performer. Moreover, some of the regularization techniques outperform both Performer and vanilla Transformer. 

\Figure[t!](topskip=0pt, botskip=0pt, midskip=0pt)[width=0.99\columnwidth]{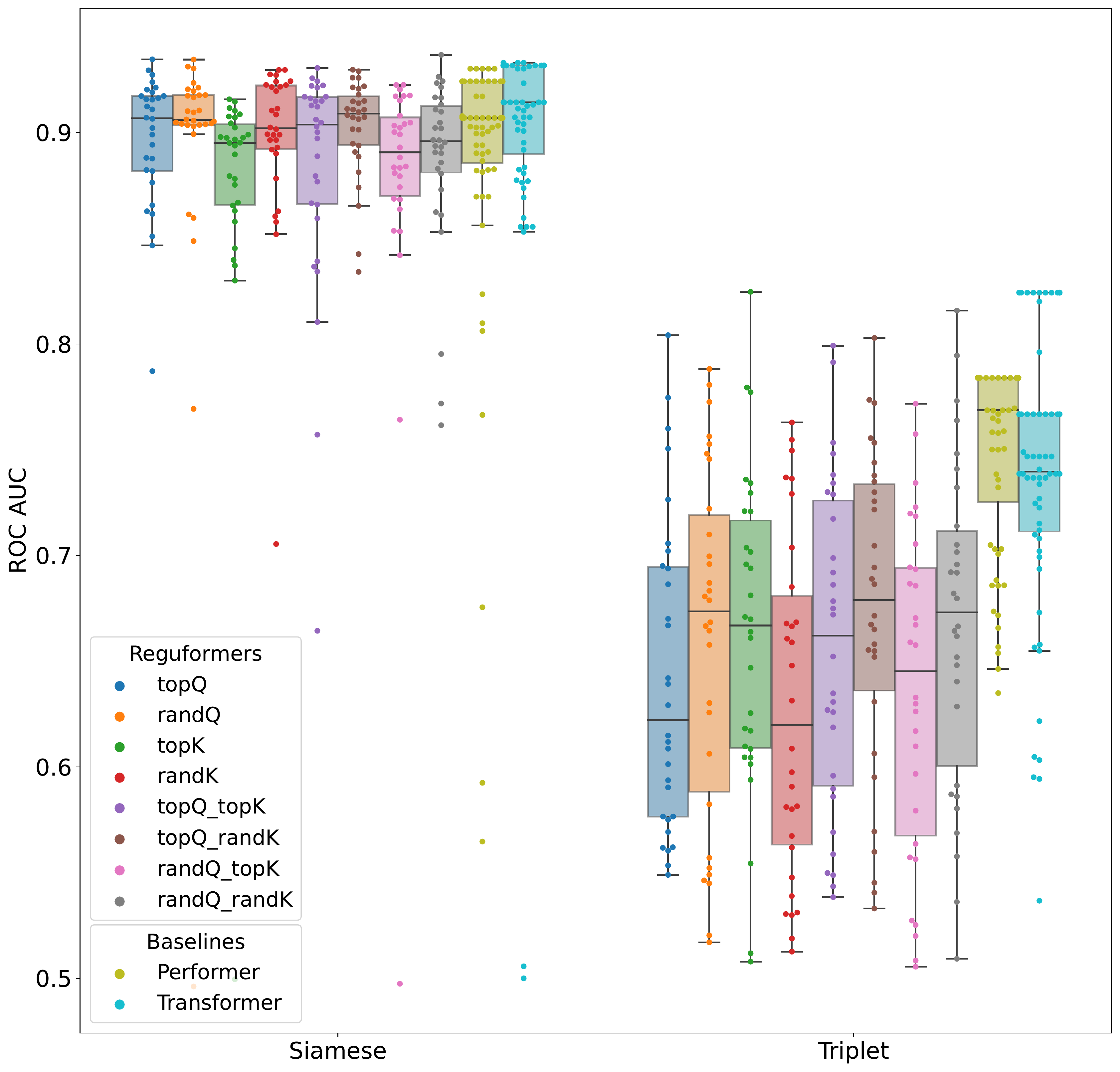}
{ \textbf{ROC~AUC scores of Transformer, Regularized Transformer (Reguformer), and Performer in Siamese and Triplet architectures during hyperparameter optimization.}\label{fig:box_plots_roc_auc}}

We evaluate the models' quality via cross-validation with five splits.
For each split, we train our model with $25000$ pairs in the case of Siamese-based approach or triplets in the case of Triplet-based approach.
For testing, we use $5000$ pairs or triplets from other wells. 

According to PR~AUC values presented in Table~\ref{tab:well_linking} for the models with the best hyperparameters, the usage of Transformer-based architectures improves the previous results obtained with RNN. 
Moreover, the scores of Reguformers are close to the results of the classical Transformer model. This experiment also shows that there is no need to use only top queries as the Informer model; random queries or keys are enough to reach the acceptable quality. In addition, the best results among all the Siamese and Triplet configurations belong to the regularized keys matrix and the combination of regularization query and key matrices simultaneously. These models achieve the PR~AUC scores comparable to the classical Transformer. 

\begin{table*}
    \centering
    \caption{\textbf{Comparison of the quality of models for well-linking problem. TOP-1 values are highlighted with \textbf{bold} font, and TOP-2 best values are \underline{underlined}. The LSTM reference results are from \cite{romanenkova2022similarity}; all other results are from our experiments}}
    \label{tab:well_linking}
        \begin{tabular}{lccccc}
        
        \hline
\multirow{2}{*}{Model} &\multicolumn{5}{c}{PR AUC} \\ \cline{2-6}
&Siam. &Siam.+Eucl.dist. &Siam.+Cos.dist. &Tripl.+Eucl.dist. &Tripl.+Cos.dist. \\ \hline
LSTM~\cite{romanenkova2022similarity} &0.951 ± 0.041 &0.894 ± 0.071 &0.930 ± 0.051 &0.909 ± 0.048 &0.903 ± 0.046 \\ 
Transformer &0.958 ± 0.024 &0.953 ± 0.013 &0.958 ± 0.018 &\textbf{0.984 ± 0.008} &0.974 ± 0.015 \\ 
Performer &0.97 ± 0.02 &0.938 ± 0.014 &0.917 ± 0.022 &0.93 ± 0.038 &0.874 ± 0.046 \\ 
Top Queries &\textbf{0.974 ± 0.009} &0.974 ± 0.015 &0.973 ± 0.01 &0.978 ± 0.012 &0.971 ± 0.016 \\
Random Queries &0.969 ± 0.023 &0.968 ± 0.015 &0.967 ± 0.016 &0.977 ± 0.01 &0.972 ± 0.013 \\
Top Keys &0.972 ± 0.011 &0.968 ± 0.015 &0.962 ± 0.017 &\underline{0.983 ± 0.011} &\textbf{0.978 ± 0.015} \\
Random Keys &0.97 ± 0.023 &\underline{0.978 ± 0.013} &\underline{0.977 ± 0.017} &\underline{0.983 ± 0.008} &0.972 ± 0.015 \\
Top Queries Top Keys &0.965 ± 0.024 &0.97 ± 0.016 &0.965 ± 0.018 &0.981 ± 0.007 &0.974 ± 0.009 \\
Top Queries Random Keys &0.969 ± 0.019 &\textbf{0.979 ± 0.012} &0.973 ± 0.015 &0.98 ± 0.009 &0.974 ± 0.017 \\
Random Queries Top Keys &0.948 ± 0.052 &0.972 ± 0.014 &0.962 ± 0.025 &0.981 ± 0.008 &0.974 ± 0.016 \\
Random Queries Random Keys &\underline{0.973 ± 0.012} &0.965 ± 0.005 &\textbf{0.979 ± 0.01} &\underline{0.983 ± 0.01} &\underline{0.976 ± 0.015} \\
        \hline
        \end{tabular}
    \end{table*}

\subsection{Robustness}
\label{subsec:robustness}
The first reason for the obtained results on the well-linking problem is that Reguformers are more robust to errors in data because they use sparse attention that utilizes only part of the data and is still more robust. We conduct the experiment by replacing data with white noise and zeros to prove this statement. To analyze the models' robustness, we generate $5000$ well-intervals and analyze the dependence of the PR~AUC scores on the percentage of changed data for the models achieving top results on well-linking tasks, namely, Reguformer with random queries and keys (randQ\_randK), Reguformer with top keys (topK), compare them with the Informer model (Reguformer with top queries, topQ) and the classical Transformer. Figure~\ref{fig:robustness} demonstrates that our Reguformers' quality decreases slower in comparison with the Transformer's. Even if most of the intervals are eliminated, their performance is acceptable. 
Moreover, Reguformer modifications with random queries and keys  (randQ\_randK) and top keys (topK) are more robust than the Informer model, especially Reguformer with random queries and keys, illustrating the slowest PR~AUC decrease among the considered models. 

\Figure[t!](topskip=0pt, botskip=0pt, midskip=0pt)[width=0.99\columnwidth]{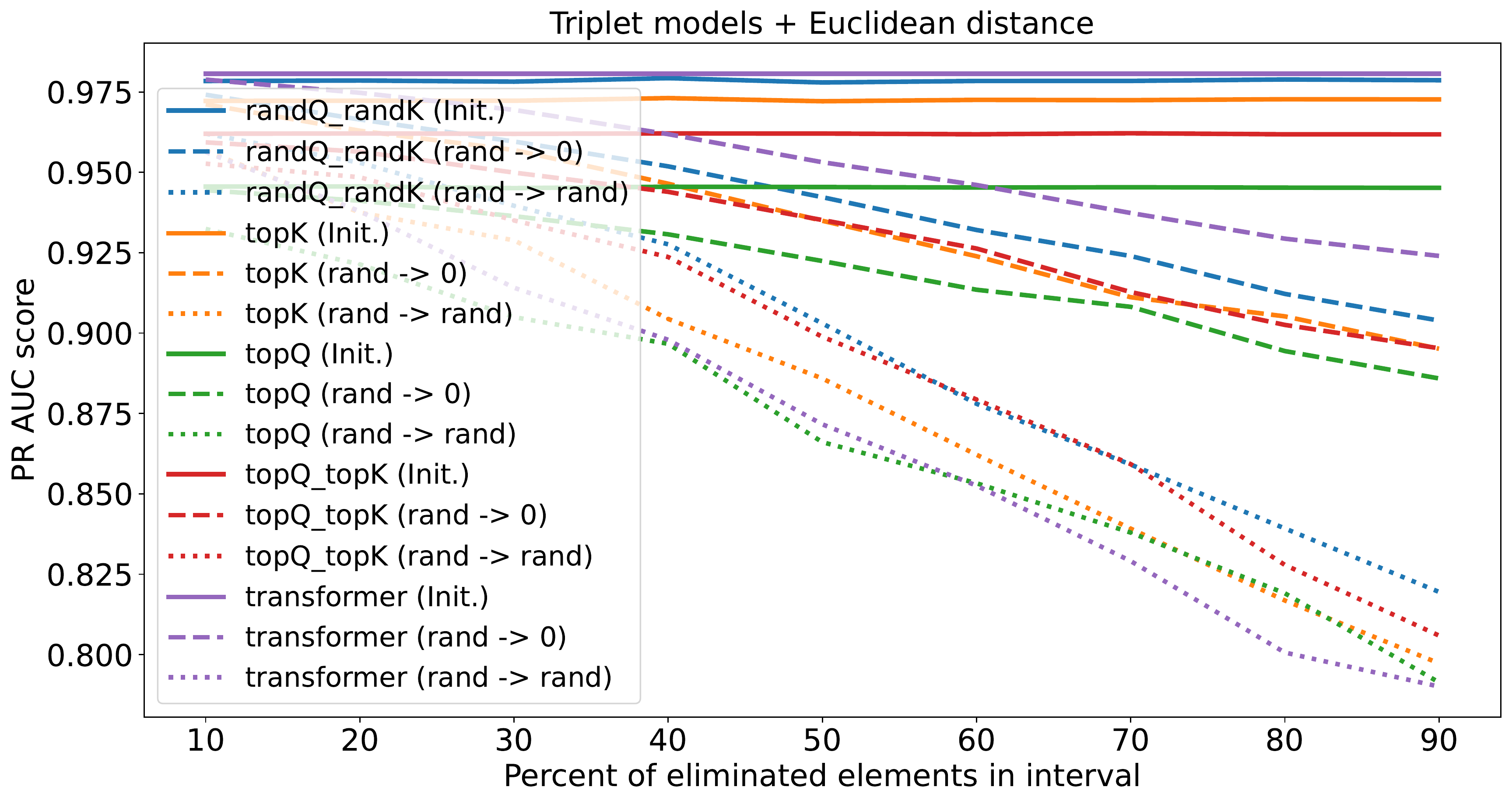}
{\textbf{The robustness of Reguformer with random queries and keys (randQ\_randK), Reguformer with top keys (topK), Reguformer with top queries (topQ) -- the Informer's analog, and the classical Transformer model. For each model, the initial PR~AUC scores are presented (Init.) and PR~AUC during the increase of eliminated random parts of well-intervals with zeros (rand -> 0) and white noise (rand -> rand).}\label{fig:robustness}}

Another reason for the well-linking task results is the data's uniform structure, with only one or two rock types in the interval. Thus, we do not need many points to make conclusions. Instead, we should be robust to errors and missing values in data. To prove this, we visualize the cosine distance between well-intervals belonging to one well and well-intervals from the two different wells. Figure~\ref{fig:one_well_cos_dist_emb} shows that parts can be easily distinguished. These changes indicate different rock types, and each rock type's data is uniform. 

\Figure[t!](topskip=0pt, botskip=0pt, midskip=0pt)[width=0.99\columnwidth]{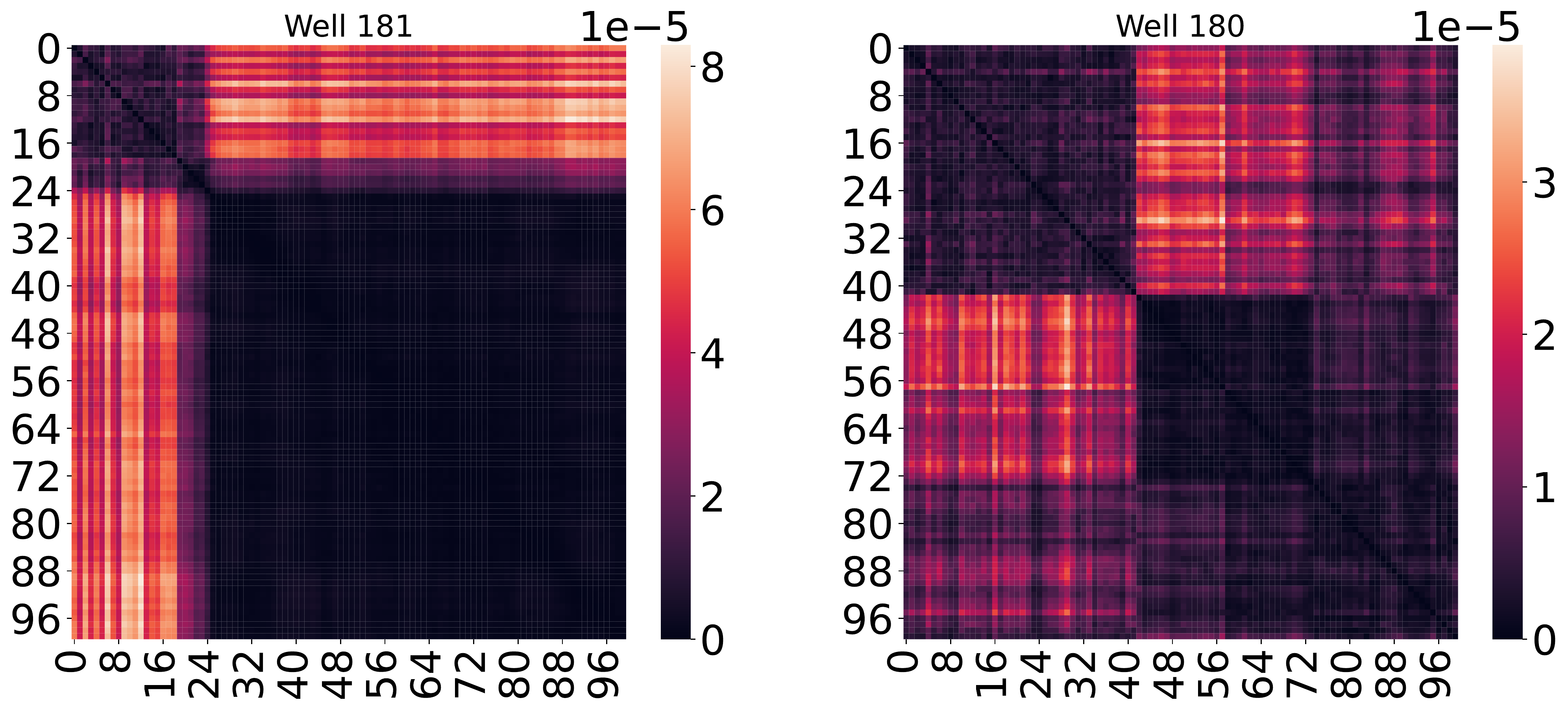}
{ \textbf{Calculated distances between intervals from one well. The two heatmaps refer to the two separate wells: well number $181$ (left) and well number $180$ (right).}
\label{fig:one_well_cos_dist_emb}}

The same effect is illustrated in Figure~\ref{fig:two_wells_cos_dist_emb} for cross-distances for intervals from different wells.
This type of heatmap can be used for rock types of two wells comparison and similarity identification between parts of wells.    

\Figure[t!](topskip=0pt, botskip=0pt, midskip=0pt)[width=0.6\columnwidth]{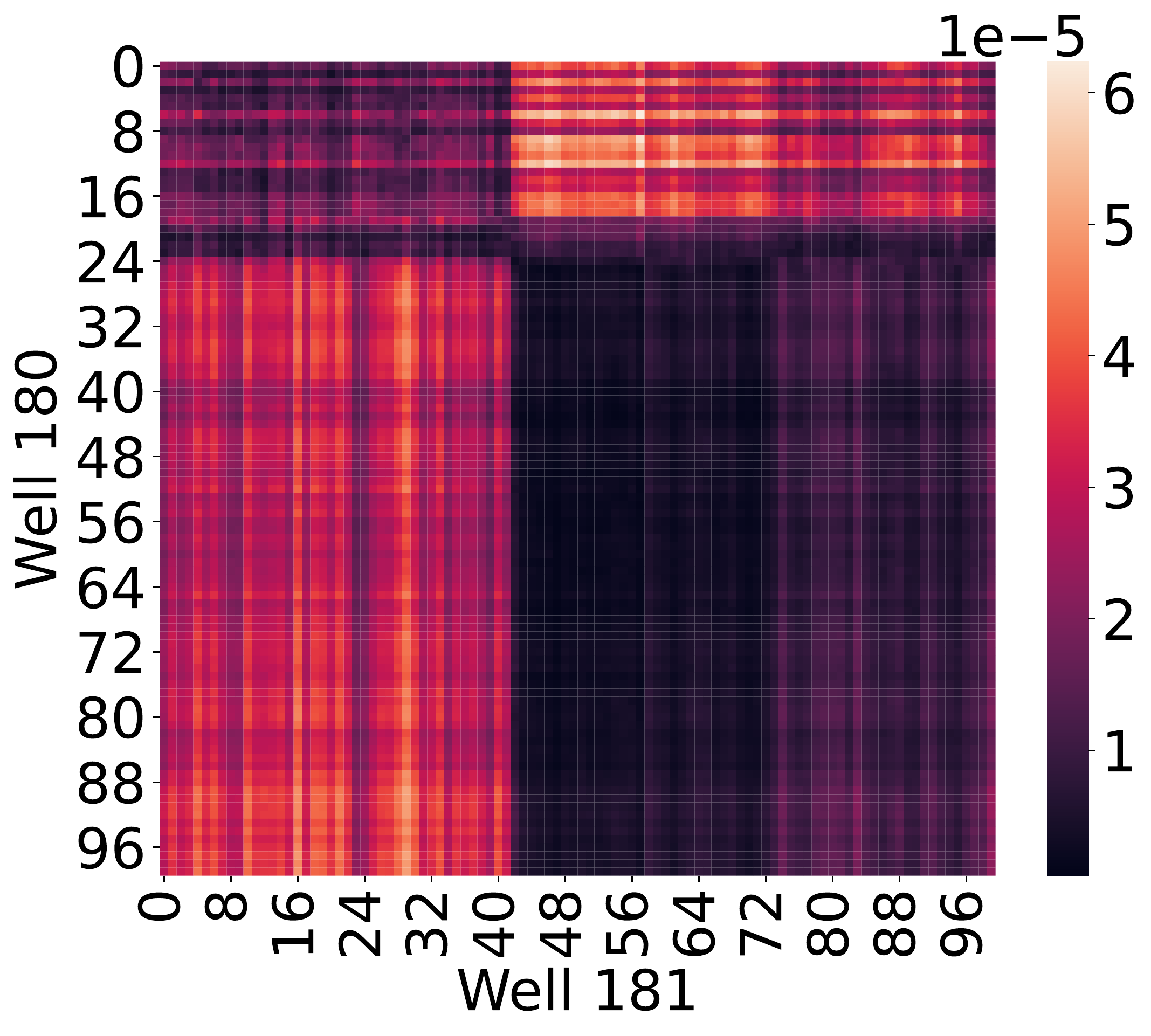}
{ \textbf{The visualization of the cosine distance between embeddings from two different wells.}\label{fig:two_wells_cos_dist_emb}}

\subsection{Embeddings quality}
\label{subsec:embeddings_quality}
After the models' training, we obtain $5000$ embeddings from models and cluster them with Agglomerative clustering. To get embeddings, we take the output of a Transformer-based encoder, flatten it, and aggregate it with a linear layer. 
The obtained clustering is compared to "true" clustering with cluster labels being well names.
All models' Adjusted Rand Index (ARI) scores are presented in Table~\ref{tab:well_linking_ari}. 
Judging by ARI, the closest score is demonstrated by Reguformer with random keys (randK). However, all other regularization techniques show a high quality of their embeddings. This experiment indicates the high quality of representations obtained via Reguformers.

\begin{table}
    \centering
    \caption{\textbf{Comparison of the embeddings' clustering quality. TOP-1 values are highlighted with \textbf{bold} font, and TOP-2 best values are \underline{underlined}. The upper part of the table contains reference results from~\cite{romanenkova2022similarity}}}
    \label{tab:well_linking_ari}
        \begin{tabular}{lcc}
        \hline
    \multirow{2}{*}{Model} &\multicolumn{2}{c}{ARI} \\ \cline{2-3}
    &Siamese & Triplet \\ 
    \hline
    XGBoost \cite{romanenkova2022similarity} &\multicolumn{2}{c}{0.258 ± 0.173}\\
    LSTM \cite{romanenkova2022similarity} &0.569 ± 0.162 &0.553 ± 0.122 \\ 
Transformer &0.793 ± 0.09 &\textbf{0.981 ± 0.037} \\ 
Performer &0.721 ± 0.09 &0.634 ± 0.058 \\
Top Queries &\underline{0.933 ± 0.076} &0.927 ± 0.06 \\
Random Queries &0.812 ± 0.122 &0.875 ± 0.105 \\
Top Keys &0.85 ± 0.117 &0.964 ± 0.061 \\
Random Keys &0.919 ± 0.053 &\underline{0.969 ± 0.062} \\
Top Queries Top Keys &0.852 ± 0.097 &0.92 ± 0.104 \\
Top Queries Random Keys &\textbf{0.948 ± 0.054} &0.941 ± 0.069 \\
Random Queries Top Keys &0.842 ± 0.079 &0.952 ± 0.057 \\
Random Queries Random Keys &0.851 ± 0.103 &0.967 ± 0.053 \\
    \hline
    \end{tabular}
\end{table}

We also try to solve a multiclass classification problem with obtained embeddings.
Class labels here are labels for particular wells.
There are $28$ of them in total in this experiment.

The procedure for the training of a downstream classifier for embeddings is the following.
We generate data intervals from the original data series (wells), which are subsequently transformed and classified. In our case, $5000$ well-intervals are generated.
        
The resulting embeddings are used as inputs to common machine learning models (downstream classifiers):
\begin{enumerate}
    \item XGBoost gradient boosting classifier with default hyperparameters;
    \item Logistic regression equivalent to one linear layer neural network;
    \item Neural network of three linear layers connected by ReLU activation function. The dimension of the first layer equals the dimension of embedding, the dimension of the second layer equals $64$, and the dimension of the third layer equals $128$. 
\end{enumerate}

Since our tasks are connected with classification, we use the following metrics: ROC~AUC, PR~AUC, and F1-score.
We present the comparison in Table~\ref{tab:emb_class}.
As we see from the table, the quality of the models is high.
This fact is true even though the classification part is simple. 
Also, observing all metrics, we can conclude that Transformer's embeddings are classified quite well, especially given that there are $28$ classes in total with the best values provided by the Siamese Reguformer with top queries (topQ) and top queries and random keys (topQ\_randK) models. However, all other regularization techniques demonstrate comparable results. 
 
\begin{table*}
    \centering
    \caption{\textbf{Mean values of quality metrics on multiclass classification task of well-intervals' embeddings, TOP-1 values are highlighted with \textbf{bold} font and TOP-2 best values are \underline{underlined}}}
    \label{tab:emb_class}
        \begin{tabular}{|c|c|c|ccc|}
        \hline
        Downstream classifier &Loss type &Regularization type &ROC AUC &PR AUC &F1 \\ \hline
\multirow{20}{*}{XGBoost} &\multirow{10}{*}{Siamese} &LSTM &0.978 &0.849 &0.814 \\
& &Transformer &0.919 &0.918 &0.951 \\
& &Top Queries &\textbf{0.997} &\textbf{0.971} &0.944 \\
& &Random Queries &\underline{0.994} &0.917 &0.875 \\
& &Top Keys &0.931 &0.929 &0.958 \\
& &Random Keys &0.911 &0.909 &0.940 \\
& &Top Queries Top Keys &0.930 &0.929 &0.959 \\
& &Top Queries Random Keys &0.959 &\underline{0.958} &\textbf{0.984} \\
& &Random Queries Top Keys &0.923 &0.922 &0.960 \\
& &Random Queries Random Keys &0.933 &0.933 &\underline{0.965} \\
&\multirow{10}{*}{Triplet} &LSTM &0.976 &0.833 &0.798 \\ \cline{2-6}
& &Transformer &0.931 &\textbf{0.931} &\textbf{0.962} \\
& &Top Queries &\underline{0.993} &0.916 &0.879 \\
& &Random Queries &\textbf{0.994} &\underline{0.926} &0.886 \\
& &Top Keys &0.895 &0.893 &0.927 \\
& &Random Keys &0.899 &0.898 &0.933 \\
& &Top Queries Top Keys &0.886 &0.887 &0.922 \\
& &Top Queries Random Keys &0.893 &0.891 &0.916 \\
& &Random Queries Top Keys &0.910 &0.909 &\underline{0.946} \\
& &Random Queries Random Keys &0.913 &0.913 &0.944 \\ \hline
\multirow{20}{*}{Linear layer} &\multirow{10}{*}{Siamese} &LSTM &0.959 &0.571 &0.578 \\
& &Transformer &0.747 &0.765 &0.766 \\
& &Top Queries &\textbf{0.997} &\underline{0.935} &\underline{0.894} \\
& &Random Queries &\underline{0.986} &0.745 &0.718 \\
& &Top Keys &0.835 &0.831 &0.875 \\
& &Random Keys &0.834 &0.835 &0.862 \\
& &Top Queries Top Keys &0.839 &0.838 &0.872 \\
& &Top Queries Random Keys &0.939 &\textbf{0.938} &\textbf{0.963} \\
& &Random Queries Top Keys &0.831 &0.825 &0.857 \\
& &Random Queries Random Keys &0.841 &0.841 &0.876 \\ \cline{2-6}
&\multirow{10}{*}{Triplet} &LSTM &\underline{0.943 } &0.528 &0.505 \\
& &Transformer &0.857 &\textbf{0.856} &\underline{0.879} \\
& &Top Queries &\textbf{0.986} &0.766 &0.735 \\
& &Random Queries &\textbf{0.986} &0.79 &0.761 \\
& &Top Keys &0.818 &0.820 &0.861 \\
& &Random Keys &0.828 &0.830 &0.869 \\
& &Top Queries Top Keys &0.793 &0.791 &0.821 \\
& &Top Queries Random Keys &0.839 &\underline{0.835} &0.876 \\
& &Random Queries Top Keys &0.786 &0.787 &0.832 \\
& &Random Queries Random Keys &0.856 &\textbf{0.856} &\textbf{0.898} \\ \hline
\multirow{20}{*}{FC NN} &\multirow{10}{*}{Siamese} &LSTM &0.979 &0.757 &0.736 \\
& &Transformer &0.867 &0.840 &0.895 \\
& &Top Queries &\textbf{0.999} &\textbf{0.972} &0.945 \\
& &Random Queries &\underline{0.994} &0.873 &0.818 \\
& &Top Keys &0.921 &0.914 &\underline{0.957} \\
& &Random Keys &0.903 &0.901 &0.935 \\
& &Top Queries Top Keys &0.923 &0.921 &0.950 \\
& &Top Queries Random Keys &0.964 &\underline{0.962} &\textbf{0.981} \\
& &Random Queries Top Keys &0.907 &0.900 &0.932 \\
& &Random Queries Random Keys &0.924 &0.920 &0.951 \\ \cline{2-6}
&\multirow{10}{*}{Triplet} &LSTM &0.972 &0.715 &0.684 \\
& &Transformer &0.914 &\textbf{0.913} &0.942 \\
& &Top Queries &\underline{0.995} &0.887 &0.834 \\
& &Random Queries &\textbf{0.996} &\underline{0.909} &0.867 \\
& &Top Keys &0.879 &0.872 &0.916 \\
& &Random Keys &0.905 &0.900 &\underline{0.944} \\
& &Top Queries Top Keys &0.855 &0.852 &0.901 \\
& &Top Queries Random Keys &0.889 &0.887 &0.927 \\
& &Random Queries Top Keys &0.900 &0.899 &0.935 \\
& &Random Queries Random Keys &0.908 &0.902 &\textbf{0.948} \\
    \hline
    \end{tabular}
\end{table*}

We also can compare the well-intervals' representations visualization. We compress $5000$ embeddings obtained via Triplet models with t-SNE  \cite{van2008visualizing} to get two coordinates for each well-interval and plot them. We consider Reguformer with random queries and keys (randQ\_randK), which shows the most promising results in terms of the well-linking problem, embeddings quality, and the model's robustness, and compare its embeddings with Triplet Transformer, LSTM from \cite{romanenkova2022similarity}, and Reguformer with top queries (topQ), which is equivalent to Informer. In Figure~\ref{fig:emb_vis}, we can distinguish the precise areas dedicated to each well for our Reguformers. This is not the case for LSTM and the classical Transformer. 
Moreover, the visualization of some wells is better if Reguformer with random queries and keys (randQ\_randK) is used.  

\Figure[t!](topskip=0pt, botskip=0pt, midskip=0pt)[width=0.99\columnwidth]{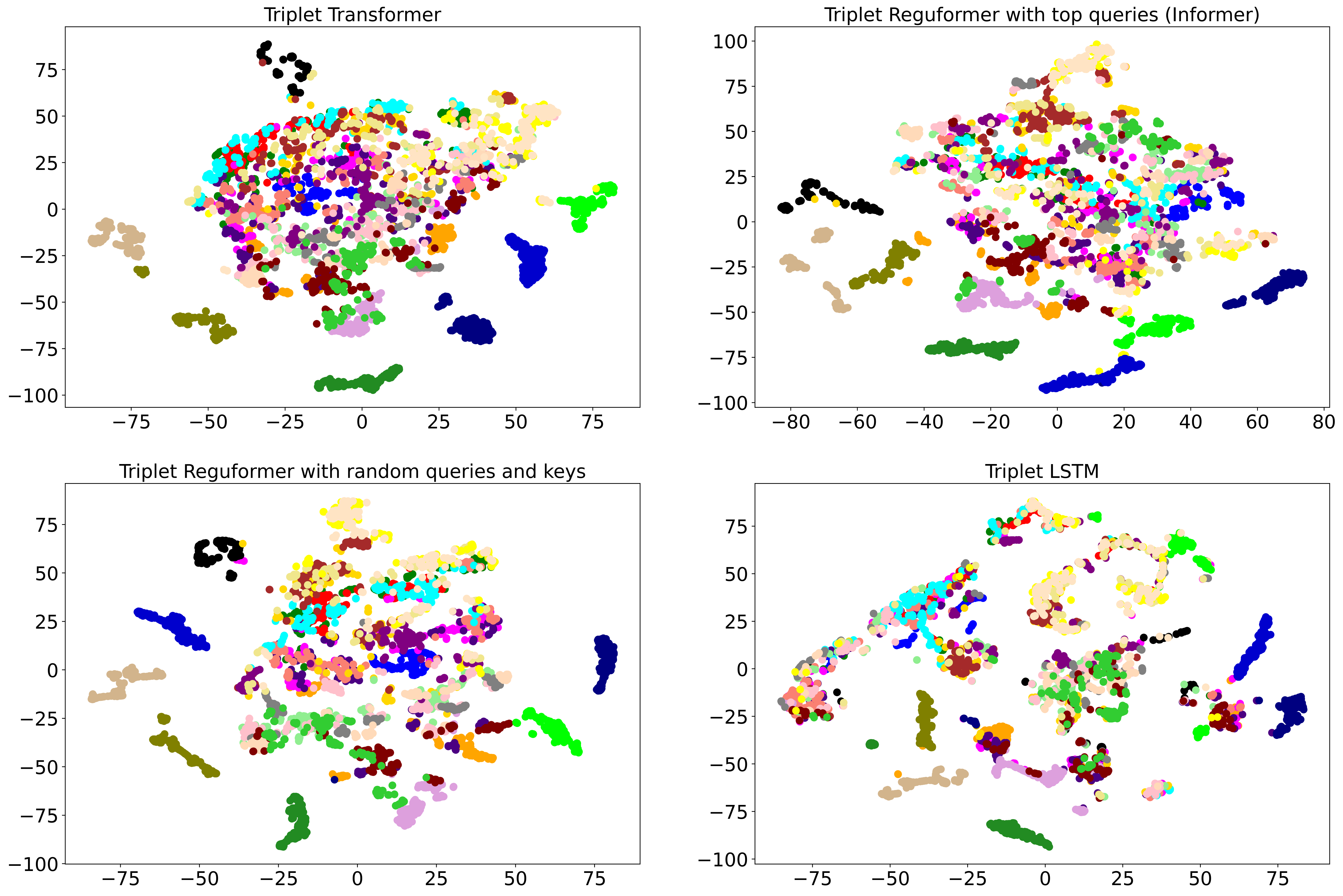}
{ \textbf{Visualization of t-SNE representation for Well-intervals embeddings. The left figure refers to embeddings obtained with Siamese Informer, while the right one -- Siamese LSTM. Each point corresponds to a single well-interval. Each well has its own color.}\label{fig:emb_vis}}


\subsection{Attention analysis}
\label{subsec:attention_analysis}

We look at the attention matrix for the pretrained Triplet Transformer with the sensitivity analysis in mind. 
We conduct two experiments:
\begin{enumerate}
    \item Estimation of correlation between Transformer attentions and models' gradients;
    \item Repeatedly replacement of a part of an interval with the lowest attention or a feature with the biggest gradient with zeros or numbers from the random normal distribution and calculate the initial and obtained model's accuracy. Each sequence element's attention is the corresponding $\mathbf{q}_i^T \mathbf{k}_i$ averaged over all layers and heads.
\end{enumerate} 

The idea is the same as the authors of \cite{kail2022scaleface} proposed: the lower attention scores, the higher uncertainty, and consequently, the highest importance of the part of an interval is. 

We generate $5000$ pairs of intervals, obtain gradients and attention weights, and calculate the correlation between them. We also examine the dependence of Siamese Transformer accuracy on the percentage of changed elements in an interval.

The correlation value is $corr = 0.02 \pm 0.08$.
Thus, there is no direction between gradients and attention values.

Although the correlation is not high enough, Figure~\ref{fig:attention_analysis} demonstrates that masking of intervals with small attention weights strongly influences the models' quality, mainly when we mask them with the elements from the normal distribution. 
We also note that Attention-based masking works differently; it is most successful when we replace given values with some random ones. The obtained results prove the statement that "Attention is not not Explanation"  \cite{wiegreffe2019attention}. 
Thus, we get additional evidence that our model is more robust to noise than others.

\Figure[t!](topskip=0pt, botskip=0pt, midskip=0pt)[width=0.99\columnwidth]{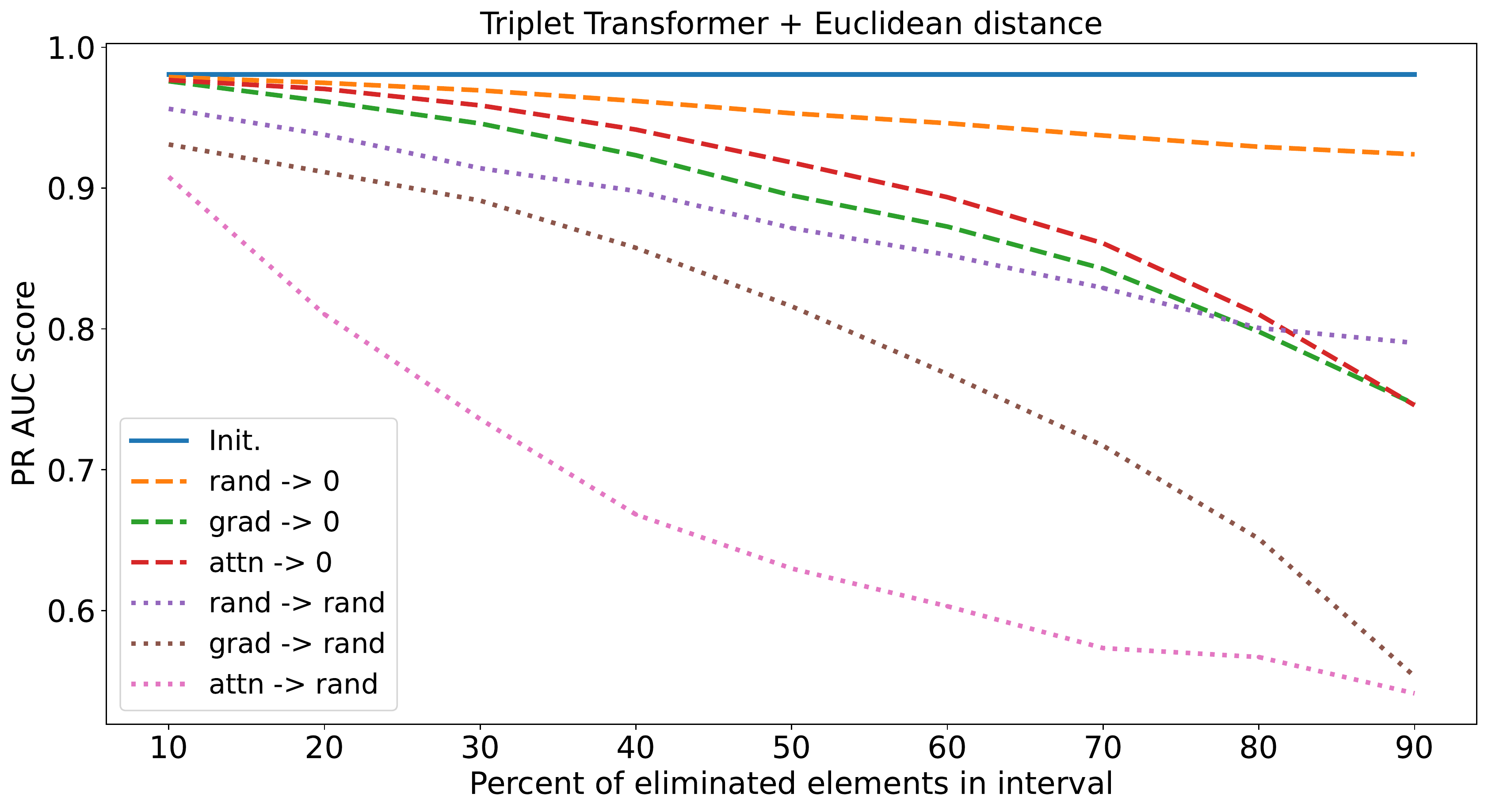}
{ \textbf{PR~AUC score during the increase of the percentage of eliminated interval's parts for different criteria for elimination. A faster decrease means that we better identify important measurements.}\label{fig:attention_analysis}}

\subsection{Inference time}
\label{subsec:inference_time}
We compare the Transformer's modifications regarding GPU inference time and the PR~AUC score. For the experiment, we used Nvidia Titan RTX GPU. Figure \ref{fig:inference_time_siamese} shows that changing the Informer's regularization strategy with top queries (topQ) to another one with random queries and keys (randQ\_randK) not only decreases the inference time but also improves the quality significantly. Some other regularizations, e.g., random queries or keys, fasten the models' performance with a small PR~AUC drop. Figure \ref{fig:inference_time_triplet} demonstrates that all previously mentioned configurations require less time for inference. However, their quality slightly decreased. In general, the PR~AUC score of all Reguformers is high. 

\Figure[t!](topskip=0pt, botskip=0pt, midskip=0pt)[width=0.99\columnwidth]{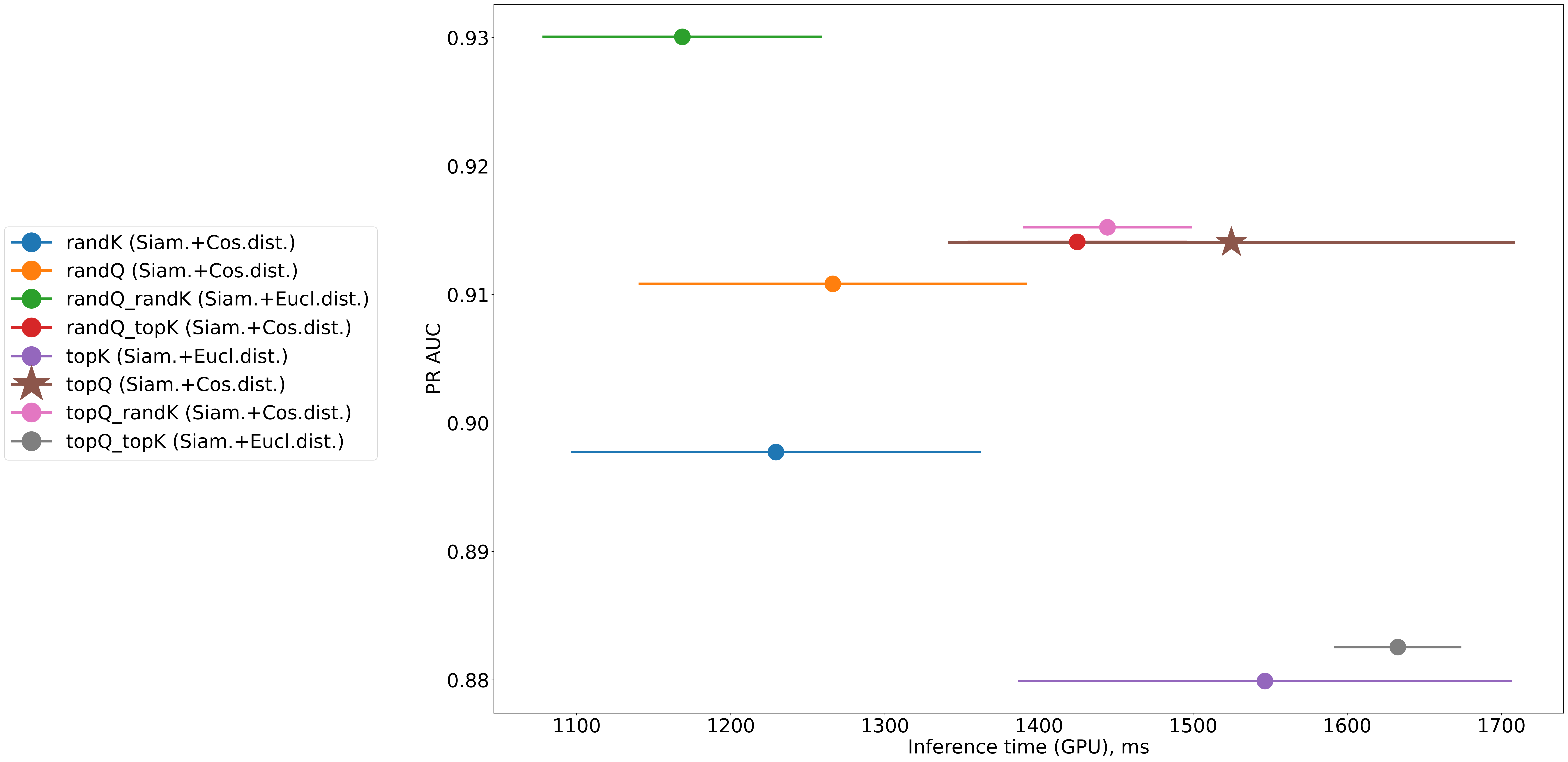}
{ \textbf{The dependence of the PR~AUC metric on the GPU inference time of Siamese models. The Informer's regularization strategy with top queries is highlighted with the star marker.}\label{fig:inference_time_siamese}}

\Figure[t!](topskip=0pt, botskip=0pt, midskip=0pt)[width=0.99\columnwidth]{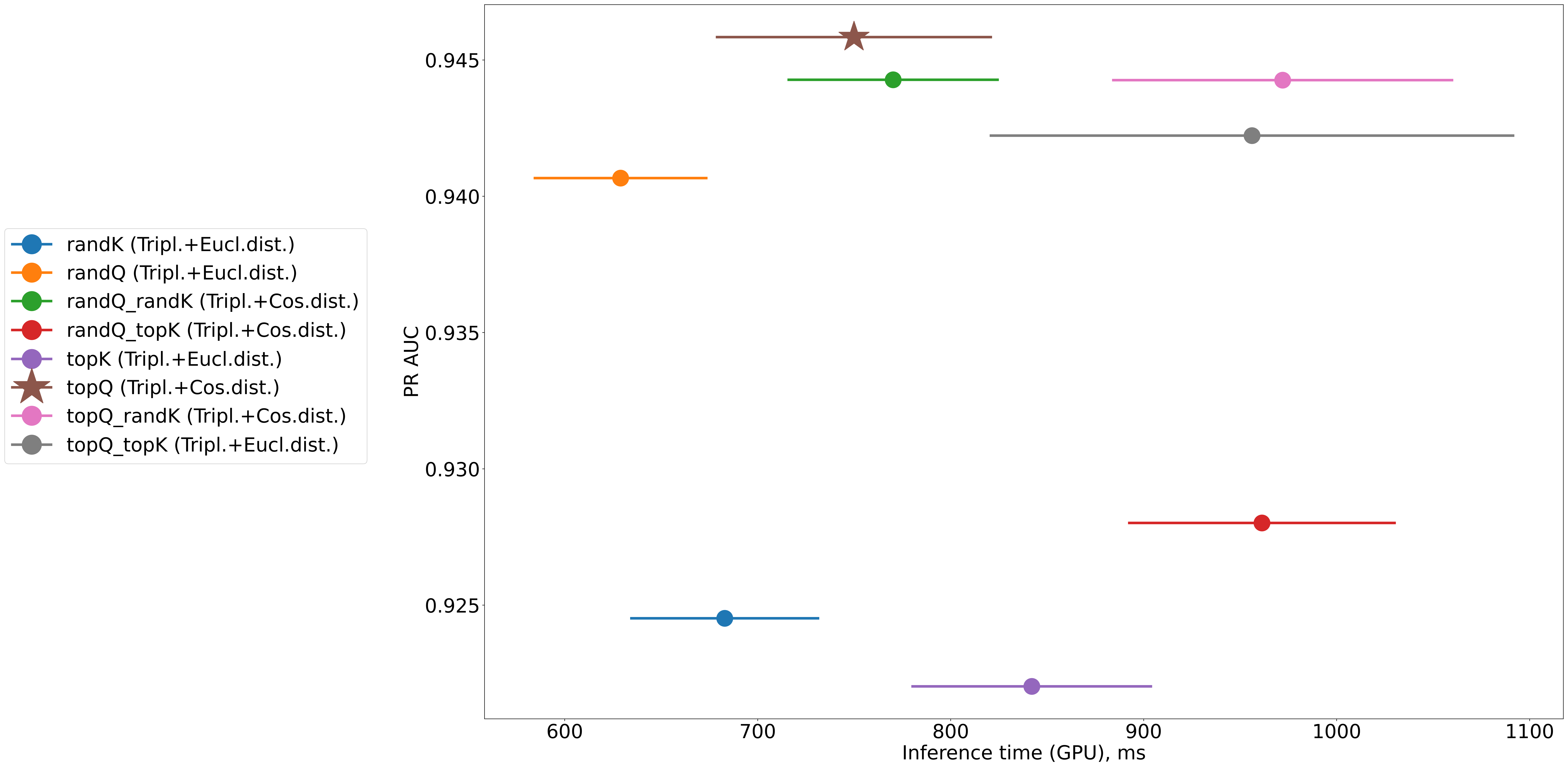}
{ \textbf{The dependence of PR~AUC on the GPU inference time of Triplet models. The Informer's regularization strategy with top queries is highlighted with the star marker.}\label{fig:inference_time_triplet}}

\section{Conclusions and discussion}
\label{sec:conclusion}

We successfully applied Transformer-based architectures to oil\&gas logging data collected during drilling. Our Reguformer models can accurately solve a similarity problem between intervals of oil wells. 
These models are better in terms of quality and efficiency, can solve diverse downstream tasks, and are more interpretable and robust than previously used.

The experimental results show an increase in quality: PR~AUC score of our Reguformer equals $0.983$, which is almost comparable to the quality of the Vanilla Transformer $0.984$, while previous models give only $0.951$.
We note that such improvement is in place for several variations of the Reguformer, namely, with top keys, random keys, and random queries and keys. The variant of Informer (Reguformer with top queries) achieves PR~AUC equal to $0.978$. 
Other variations of Reguformers provide quality better than previously used LSTM-based similarity estimation model and are similar to Informer. 
Judging by embeddings in the clustering experiment, our Reguformers, both in Siamese and Triplet variations, provide meaningful well-interval representations in terms of geological scope. Our best ARI score is equal to $0.969$ for the random keys regularization strategy, which surpasses Informer with ARI equal to $0.933$.
It is again higher than the previously obtained value of $0.569$ and is closer to the perfect one of $1$. The ARI score for the original Transformer is not significantly higher, being $0.981$. 

Our conclusions about strong embeddings are also exposed by the experiment with embeddings' classification on wells by a simple machine learning algorithm on top of embeddings. This experiment demonstrates the higher quality of embeddings of Reguformers than the Vanilla Transformer.   
Moreover, we show that attention maps in our models provided their interpretability. 
For our model, lower attention values correspond to more important parts of an interval: a model quality decreases with increasing the percentage of eliminated parts of intervals by a low attention score.
Furthermore, we show the decrease in inference time during the best regularization strategies mentioned above: top keys, random keys, and random queries and keys. All these techniques allow to fasten the models' performance and even significantly increase the model's quality in the case of the Reguformer with random queries and keys, especially in Siamese configuration.



\bibliographystyle{IEEEtran}
\bibliography{bibliography}

\appendices
\section{Data preprocessing}
The data preprocessing strategy resembles that of  \cite{romanenkova2022similarity} and is the following: 
\begin{enumerate}
    \item Eliminate intervals with negative or zero-equal resistivity and cavernous intervals with the difference between calliper and bit size is greater than $0.35$.
    \item Fill missing values via forward and backward fill.
    \item Convert all electrical resistivity data to log-normal scale.
    \item Normalize gamma-ray and neutron log data within each well and formation using the standard scaler.
    \item Normalize other features by subtracting the mean and dividing by unit variance.
\end{enumerate}

\section{Technical details}
\label{sec:tech_details}

\begin{table}
    \caption{\textbf{The best hyperparameters and hyperparameters' search spaces for Siamese- and Triplet- Transformer, Informer, and Performer. We use curly brackets for denoting sets, Python notation for ranges: start value, excluded stop value and step, and $[\text{minimum}, \text{maximum}]$ notation for float features optimization in Optuna}}
    \label{tab:best_params}
    \begin{tabular}{|c|cc|c|}
        \hline
        Parameters & Siamese & Triplet & Search Space \\
        \hline
        \multicolumn{4}{c}{Transformer} \\
        \hline
        \texttt{num\_heads} & 8 & 6 & $\{2, 4, 6, 8\}$ \\ 
        \texttt{dropout\_prob} & 0.156 & 0.398 & [0.1, 0.5] \\ 
        \texttt{dim\_model} &32 &16 & $\{16, 32, 64\}$ \\ 
        \texttt{dim\_feedforward} & 128 & 1024 & $\{128, 512, 1024\}$ \\ 
        \texttt{num\_layers} & 3 &5 &range(2, 7, 1) \\ 
        \hline
        \multicolumn{4}{c}{Informer} \\
        \hline
        \texttt{num\_heads} & 8 & 6 & $\{2, 4, 6, 8\}$ \\ 
        \texttt{dropout\_prob} & 0.156 & 0.398 & [0.1, 0.5] \\ 
        \texttt{dim\_model} &32 &16 & $\{16, 32, 64\}$ \\ 
        \texttt{factor} & 5 & 11 & range(3, 12, 2) \\ 
        \texttt{dim\_feedforward} & 128 & 1024 & $\{128, 512, 1024\}$ \\ 
        \texttt{num\_layers} & 3 &5 &range(2, 7, 1) \\ 
        \hline
        \multicolumn{4}{c}{Performer} \\
        \hline
        \texttt{num\_heads} &4 &2 & $\{2, 4\}$ \\ 
        \texttt{dropout\_prob} &0.101 &0.186 &[0.1, 0.9] \\ 
        \texttt{num\_rand\_features} &3 &4 &range(1, 5, 1) \\
        \hline
    \end{tabular}
\end{table}

For Transformer and Informer we use our code that is available at \texttt{roguLINA/Reguformer} \footnote{\underline{https://github.com/roguLINA/Reguformer}}. It is inspired by \texttt{zhouhaoyi/Informer2020} \footnote{\underline{https://github.com/zhouhaoyi/Informer2020}}. For Performer -- \texttt{nawnoes/pytorch-performer} \footnote{\underline{https://github.com/nawnoes/pytorch-performer}}.

For hyperparameters optimization for all models, we use Optuna \footnote{\underline{https://github.com/optuna/optuna}} with $30$ iterations, group $2$-fold cross-validation with $20$ epochs for each split and each model. Batch size equals to $64$. 

We vary the most essential hyperparameters for \textit{all} models: 
\begin{itemize}
    \item \texttt{num\_heads} -- the number of heads for multi-head attention;
    \item \texttt{dropout\_prob} -- the number of heads for multi-head attention, and the dropout probability.
\end{itemize}

In addition, we vary:

\begin{itemize}
    \item for vanilla Transformer:
    \begin{enumerate}
        \item \texttt{dim\_model} -- the model embedding dimension,
        \item \texttt{dim\_feedforward} -- the dimension of the fully-connected layers,
        \item \texttt{num\_layers} -- the number of layers in the encoder.
    \end{enumerate}
    \item for Informer:
    \begin{enumerate}
        \item \texttt{dim\_model} -- the model embedding dimension,
        \item \texttt{factor} -- the probsparse attention factor,
        \item \texttt{dim\_feedforward} -- the dimension of fully-connected network part,
        \item \texttt{num\_layers} -- the number of encoder layers.
    \end{enumerate}
    \item for Performer:
    \begin{enumerate}
        \item \texttt{num\_rand\_features} -- the number of random features.
    \end{enumerate}
\end{itemize}

We use $\texttt{hidden\_size} = 64$ in FC part of Siamese architecture. The best hyperparameters and hyperparameters' search space for each model are presented in Table~\ref{tab:best_params}. 

All these models with the best hyperparameters were tested via group $5$-fold cross-validation. As it can be seen from Figure \ref{fig:box_plots_roc_auc}, our Reformers outperform the classical Transformer.

\begin{IEEEbiography}[{\includegraphics[width=1in,height=1.25in,clip,keepaspectratio]{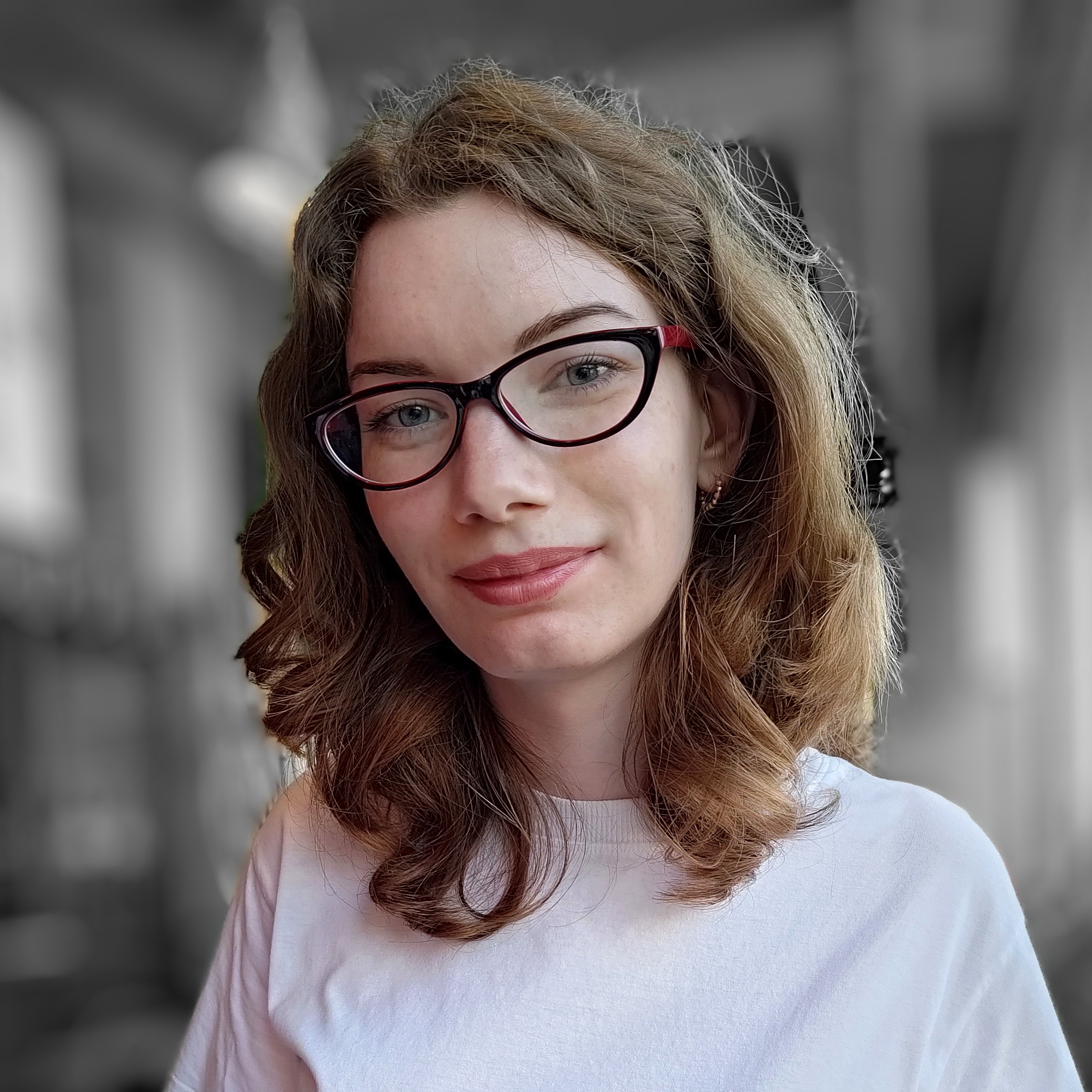}}]{Alina Ermilova} 
received the B.S. degree in Information Science and Computation Technology from
the National Research University – Higher School of Economics (HSE University), Moscow, in 2021 and the M.S. degree in
Data Science from The Skolkovo Institute of Science and Technology (Skoltech), Moscow, in 2023.
Since 2020, she has been working at Applied AI Center of Skoltech.
Her research interests include time-series processing, efficient transformers, models robustness, adversarial attacks.   
\end{IEEEbiography}

\begin{IEEEbiography}
    [{\includegraphics[width=1in,height=1.25in,clip,keepaspectratio]{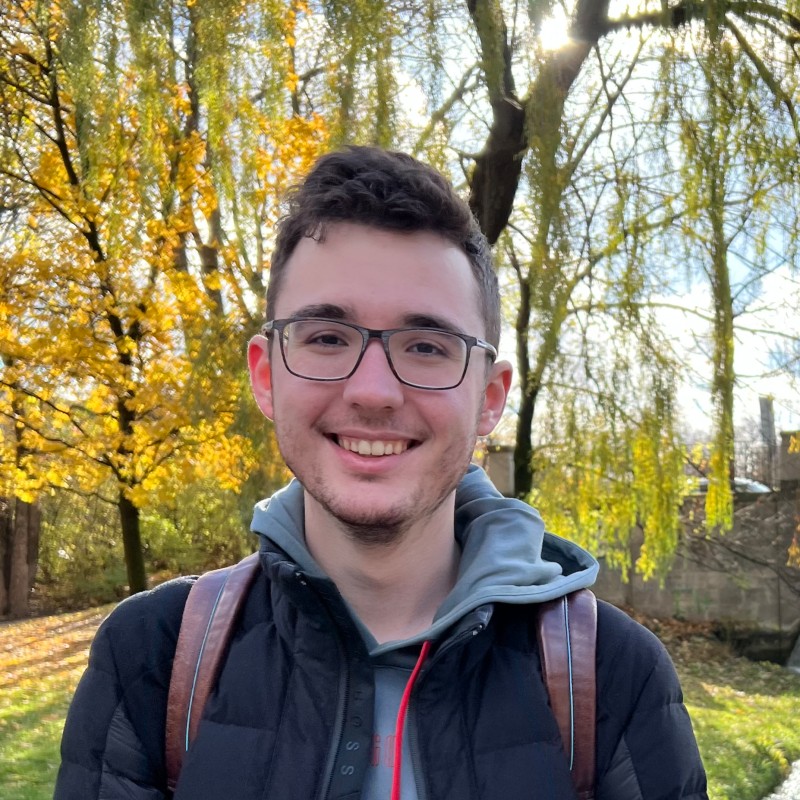}}]{Nikita Baramiia} received the B.S. degree in Economics from the Economic Faculty of the Lomonosov Moscow State University, Moscow, in 2021 and the M.S. degree in Data Science from The Skolkovo Institute of Science and Technology (Skoltech), Moscow, in 2023. Since 2020 he has gained diverse experience from working in jewelry and bank industries to classifieds IT company Avito as well as research activity in Skoltech.  His research interests include self-supervised learning, models robustness, cost-effective adversarial training.
\end{IEEEbiography}

\begin{IEEEbiography}
    [{\includegraphics[width=1in,height=1.25in,clip,keepaspectratio]{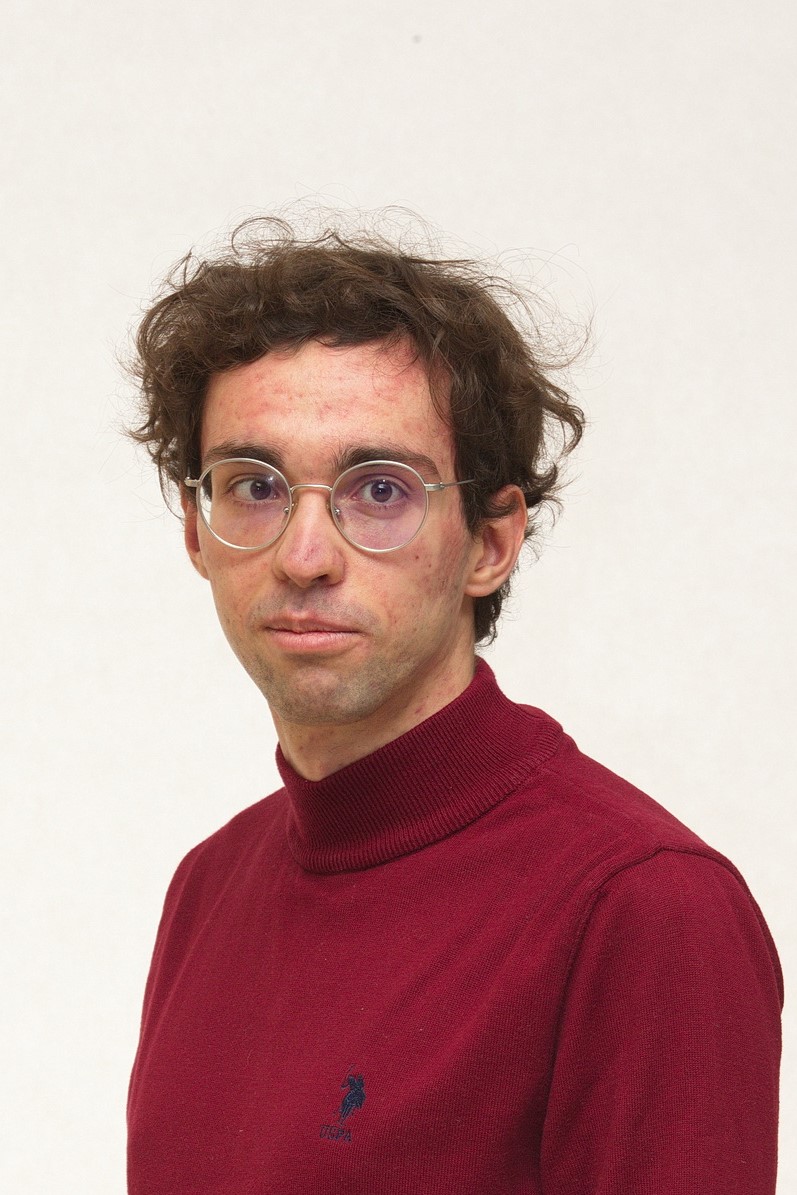}}]{Valerii Kornilov} received the B.S. degree in Economics from the Economic Faculty of the Lomonosov Moscow State University, Moscow, in 2021 and the M.S. degree in Data Science from The Skolkovo Institute of Science and Technology (Skoltech), Moscow, in 2023. His research interests include classic ML and DL approaches for robotics tasks. The primary topic of interest is foundational models for manipulation and navigation.
\end{IEEEbiography}

\begin{IEEEbiography}
    [{\includegraphics[width=1in,height=1.25in,clip,keepaspectratio]{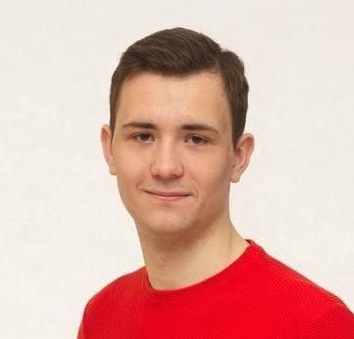}}]{Sergey Petrakov} received the B.S. degree in Economics from the Faculty of Economics, Lomonosov Moscow State University, Moscow, Russia in 2021 and the M.S. degree in Data Science from Skoltech, Moscow, Russia in 2023. He is currently pursuing the Ph.D. degree in Computational and Data Science and Engineering at Skoltech. His research interests include application of uncertainty estimation methods to NLP tasks, modern transformer-based models.
\end{IEEEbiography}

\begin{IEEEbiography}[{\includegraphics[width=1in,height=1.25in,clip,keepaspectratio]{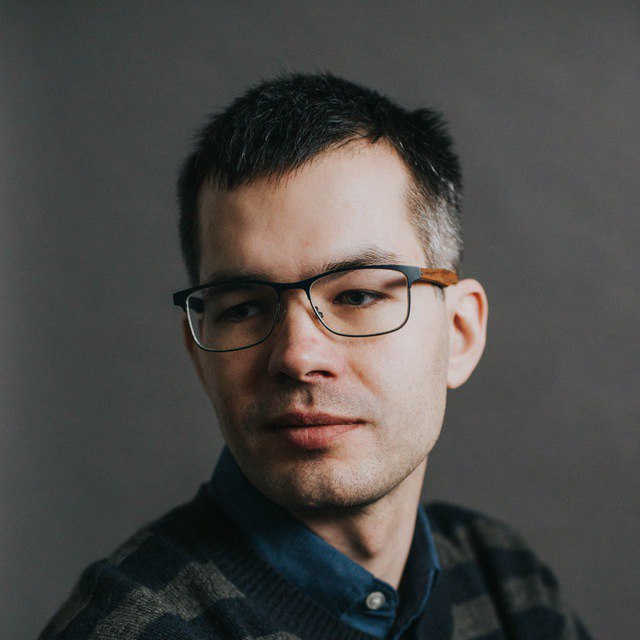}}]{Alexey Zaytsev} was born in Kharkiv, Ukraine.
He received the graduate degree from MIPT,
in 2012, and the Ph.D. degree in mathematics
from IITP RAS, in 2017. In his master’s thesis,
he proposed a modification of the Bayesian
approach for linear regression that allows an
automated feature selection. He is currently an
Assistant Professor at Skoltech. His research
interests include the development of new methods
for sequential data, Bayesian optimization, and
embeddings for weakly structured data.
\end{IEEEbiography}

\EOD

\end{document}